\documentclass{article}

\usepackage[protrusion=false]{microtype}
\usepackage{microtype}      
\usepackage[table]{xcolor}  
\usepackage{amsfonts}       
\usepackage{tcolorbox}      
\usepackage{arxiv}
\usepackage{verbatim} 
\usepackage{array}
\usepackage{hyperref}       
\usepackage{url}            
\usepackage{booktabs}       
\usepackage{amsfonts}       
\usepackage{nicefrac}       
\usepackage{microtype}      
\usepackage{lipsum}		
\usepackage{graphicx}
\usepackage{natbib}
\usepackage{doi}
\usepackage{amsmath}
\usepackage{listings}

\usepackage{xcolor}

\title{MMTClinic: Multimodal, Multilingual Time Series Question Answering and Reasoning Benchmark for Clinical Domain}

\author{
Sourav Malakar\\
Department of CSE\\
Institute of Engineering and Management\\
Kolkata, West Bengal, India\\
\texttt{sourav.malakar@iem.edu.in}
\And
Harshit Nigam\\
Department of CSE\\
IIT Patna\\
Bihar, India
\And
Akash Ghosh\\
Department of CSE\\
IIT Patna\\
Bihar, India
\And
Sriparna Saha\\
Department of CSE\\
IIT Patna\\
Bihar, India
\And
Amlan Chakrabarti\\
Department of A. K. Choudhury School of IT\\
University of Calcutta\\
Kolkata, West Bengal, India
\And
Saptarsi Goswami\\
Department of Computer Science\\
Bangabasi Morning College\\
Kolkata, West Bengal, India
\And
Dr. Priti Singh\\
Department of CSE\\
IIT Patna\\
Bihar, India
}

\renewcommand{\shorttitle}{\textit{arXiv} Template}

\hypersetup{
pdftitle={A template for the arxiv style},
pdfsubject={q-bio.NC, q-bio.QM},
pdfauthor={David S.~Hippocampus, Elias D.~Striatum},
pdfkeywords={First keyword, Second keyword, More},
}

\begin{document}
\maketitle

\begin{abstract}
	Time-series data in clinical settings is crucial for capturing dynamic changes in a patient's health over time, enabling timely diagnosis, personalized treatment, and early detection of critical events. However, the development of clinically reliable and linguistically inclusive medical AI systems remains a significant challenge, primarily due to the lack of multimodal, multilingual, and time-series-grounded benchmarks that reflect the complexity of real-world clinical scenarios. To fill this gap, we present MMTClinic, a benchmark designed to evaluate large language models (LLMs) on complex reasoning and question-answering tasks involving clinical time-series. MMTClinic combines text, medical images, and multivariate physiological signals and includes 30,000 QA pairs (15,000 multiple choice questions (MCQs) and 15,000 open-ended questions) across five languages: English, Hindi, Bengali, Marathi, and Tamil. These questions cover three important clinical tasks—mortality prediction, heart rate forecasting, and SOFA score estimation. We evaluate 13 state-of-the-art LLMs in zero-shot, few-shot, and chain-of-thought settings. Our evaluation reveals notable differences in model performance across tasks, languages, and modalities, highlighting current limitations in clinical reasoning capabilities. MMTClinic provides a valuable resource for advancing multilingual, multimodal, and time-series-aware medical AI research. The dataset will be made publicly available on successful acceptance of the work.
\end{abstract}


\section{Introduction}
Large language models (LLMs) and Multimodal Large Language Models (MLLMs) have demonstrated remarkable performance in a variety of domains, including healthcare\cite{chan2024medtsllm,hu2023nurvid,tu2024towards,ye2021unified}. State-of-the-art models, such as GPT \cite{openai2023gpt4}, are progressively gaining popularity to support clinical decision making in the healthcare domain. 
However, the majority of studies in the healthcare domain have primarily focused on text or text--image combinations, leaving a notable gap in addressing more complex modalities such as time series data.\par

The integration of time series, textual, and visual data is critical for understanding complex real-world phenomena, where numerical trends and contextual narratives jointly facilitate comprehensive analysis and decision-making. Although significant work exists in domains such as finance \cite{jacob1999fintime,hu2025fintsb} and weather forecasting \cite{rasp2020weatherbench} illustrate the need for models that can jointly interpret numerical and textual data; however, this capability remains mostly underexplored in the healthcare domain. \par

\textbf{Research Gap.}
Existing clinical time-series reasoning benchmarks fall into three categories: (i) signal-only MCQs \cite{oh2023ecgqa,wang2025ecgexpertqa,pham2025qheart}, which lack clinical text and imaging; (ii) text + time-series datasets \cite{kim2025timer,kweon2024ehrnoteqa}, which omit visual data and MCQ formats; and (iii) comprehensive multimodal datasets \cite{bae2023ehrxqa,hayat2022medfuse}, which include text, signals, and images but rely on open-ended tasks and static radiographs. Specifically, none supports multilingual MCQ-based-reasoning with integrated clinical text, time-series, and images. \textbf{\textit{To date, no benchmark unifies clinical text, images, and time-series data in an MCQ and reasoning-based framework in multiple languages.}}

\textbf{Present Work:} To address the limitations of existing benchmarks, we propose \textbf{\textit{MMTClinic}} a \textbf{M}ultimodal and \textbf{M}ultilingual \textbf{T}ime-series question-answering and reasoning benchmark tailored for the healthcare domain. \textbf{\textit{MMTClinic}} supports two complex downstream tasks: \textbf{(a)} multiple-choice question answering (MCQ-QA), and \textbf{(b)} open-ended clinical decision-making (Reasoning).  Unlike previous efforts, our benchmark integrates multivariate clinical time-series data with both textual and visual modalities to capture and reason over temporal trends, critical for accurate disease progression analysis. The dataset is constructed using a semi-automated pipeline based on the 2012 PhysioNet/CinC Challenge dataset~\cite{silva2012predicting}, from which we generate a total of 15k MCQ and reasoning QA pairs, in total 30k samples. These samples are crafted across five linguistically diverse Indian languages: \textbf {English, Bengali, Hindi, Marathi, and Tamil.}\par
\begin{figure}[!htbp]
\centering
  \includegraphics[scale=0.25]{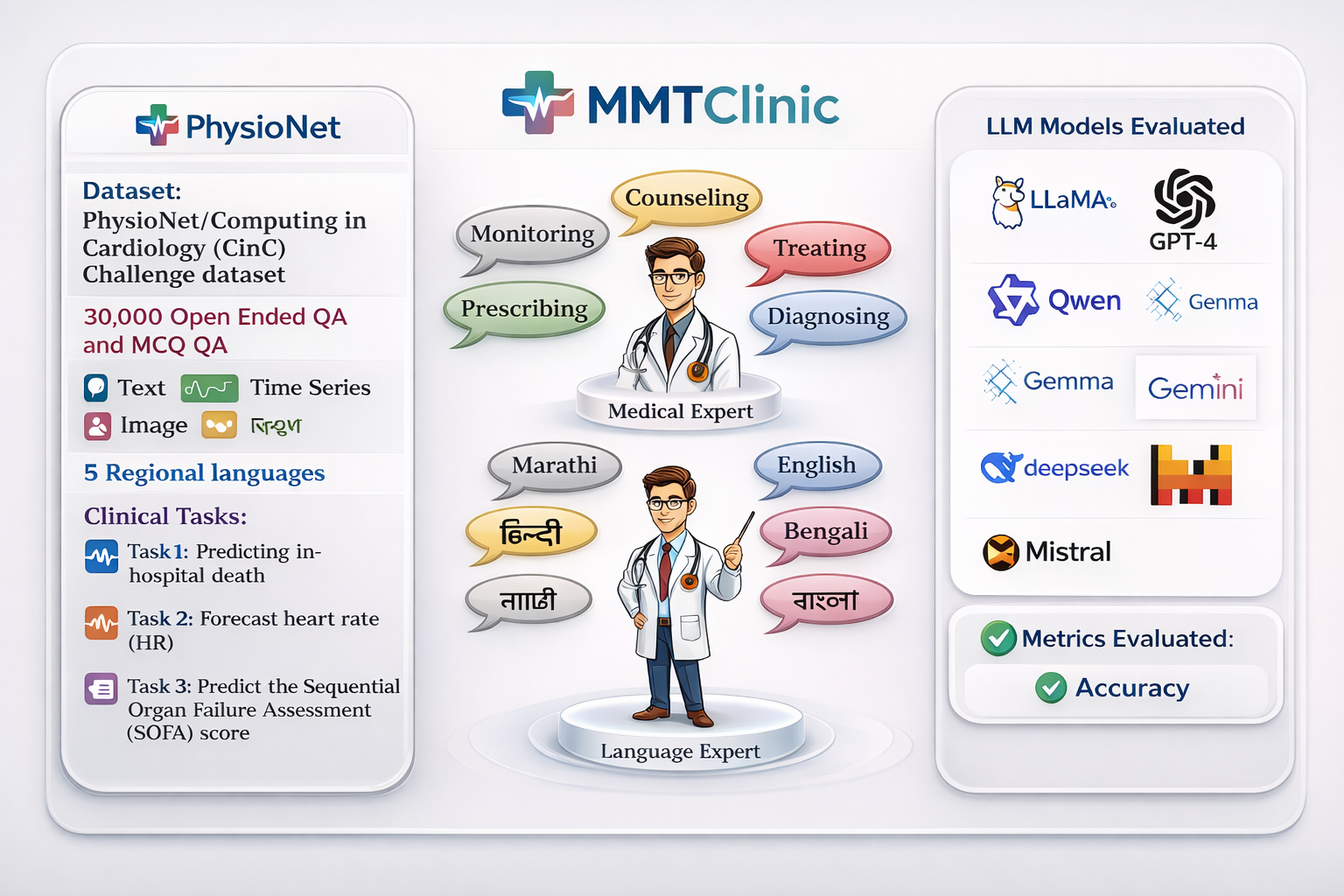}
  \caption{Overview of MMTClinic   }
  \label{fig:mmt1}
\end{figure}

We have evaluated a diverse set of \textbf{13 language models}, encompassing \textbf{LLMs}--both proprietary and open-source—along with \textbf{Small Language Models (SLMs)}, \textbf{Reasoning Language Models (RLMs)} and \textbf{Multimodal LLMs (MLLMs)}. This broad evaluation setup enables a linguistically rich and modality-aware assessment across both MCQ and open-ended reasoning tasks involving text, images, and time-series data. The benchmark includes \textbf{three real-world ICU-centered clinical tasks}: \textbf{(1)} Predicting in-hospital mortality (binary classification), \textbf{(2)} Forecasting heart rate trends using the first 48 hours of ICU time-series data, and \textbf{(3)} Estimating the SOFA score from 48-hour multivariate physiological signals. Figures  \ref{fig:mmt1} illustrate the multimodal structure and overall design of \textbf{MMTClinic}.

\textbf{Contributions:}

In summary, the main contributions of the \textbf{\textit{MMTClinic}} study are as follows.

\textbf{Comprehensive Evaluation:}  
We perform joint reasoning over text, images, and time-series across two QA formats, four modality settings, three ICU tasks, and three prompting modes.

\textbf{Domain Specific Healthcare Coverage:}  
The benchmark targets early ICU care (first 48 hours), capturing six vital signs in both numeric (.csv) and visual (.png) formats.

\textbf{Multilingual Coverage:}  
We release 30,000 QA pairs (15k MCQ, 15k reasoning) in five languages: English, Hindi, Bengali, Marathi, and Tamil for fair multilingual evaluation.

\textbf{Diverse Model Benchmarking:}  
13 models, including general LLMs, reasoning-optimized LRMs, and MLLMs are benchmarked under all prompt settings, with analysis across tasks, languages, and modalities.

\textbf{Expert-Curated Quality:}  
Data were generated semi-automatically and carefully refined by physicians and linguists. Specifically, \textbf{approximately 70\% of QA samples from each task and modality were independently reviewed and scored by four certified medical experts} for clinical accuracy and reasoning quality. Additionally, \textbf{native-speaking linguists checked all of the multilingual translations}. Samples underwent quality control through expert scoring (shown in Table~\ref{tab:expert_ratings_clean}), targeted post-editing, and reasoning consistency checks.


\par

\section{Related Works}
\label{rw}
In recent years, LLMs have been extensively used across diverse domains such as
healthcare, education, law, finance \cite{mariani2026evaluating} and scientific research, demonstrating remarkable
adaptability and performance \cite{brown2020language}. In spite of the growing
impact of LLMs across various domains, their application in clinical time series
analysis remains notably limited. Most existing LLMs are optimized for text data,
and fail to effectively handle temporal physiological signals or multivariate
time-series inputs crucial for clinical reasoning \cite{hager2024evaluation,
jin2023large, brown2025large, mousavi2023towards}.

In recent works, LLMs have mainly focused on English-centric clinical
question-answering tasks, mostly ignoring regional and low-resource languages.
This language bias raises concerns about ensuring fair access to reliable
healthcare information for different communities. Benchmarks such as MedExpQA
\cite{alonso2024medexpqa} and IndicQA \cite{singh2024indic} demonstrate that
current LLMs struggle with medical reasoning and accuracy in non-English settings.
A growing body of work has begun to target multilingual and Indic-language medical
understanding directly, including surveys of multilingual reasoning in language
models \cite{ghosh2025survey}, 
multilingual medical reasoning \cite{onyame2026cure}, trustworthiness evaluation for
healthcare LLMs across languages \cite{ghosh2025clinic}, multimodal medical query
analysis in Indian languages \cite{ghosh2026indicmedqa}, multi-agent multimodal
medical reasoning in Indic languages \cite{halder2026arogyasutra}, and summarization
of code-mixed clinical queries \cite{ghosh2024medsumm}. However, these efforts
largely address text or text-only reasoning and do not incorporate physiological
time-series signals.

Beyond the linguistic dimension, another line of research explores multimodal
clinical understanding that jointly reasons over clinical text and medical images,
such as multimodal question summarization \cite{ghosh2024clipsyntel}, clinical
document summarization \cite{ghosh2024sights, ghosh2024healthalignsumm}, and
multimodal medical retrieval \cite{acharya2025m3retrieve}. While these approaches
integrate visual and textual modalities, they rely on static images (e.g.,
radiographs) and do not model temporal trends in multivariate physiological data,
leaving multimodal temporal reasoning largely unexplored.

Recent research \cite{singhal2025toward, kim2025limitations, lievin2024can,
zhourevisiting, li2024mediq} indicates that LLMs have mostly been evaluated using
MCQ QAs in clinical tasks, which may not adequately assess their complex reasoning
capabilities. Although LLMs reveal high precision on standardized MCQ benchmarks
such as MedQA \cite{yao2024medqa} and MedMCQA \cite{pal2022medmcqa}, they often
struggle with tasks requiring nuanced clinical reasoning, such as differential
diagnosis and treatment planning.

Unlike prior work, this study unifies multivariate clinical time-series with textual
and visual modalities, introduces open-ended reasoning tasks beyond MCQs, and
evaluates a diverse set of LLMs across five languages, jointly addressing the
temporal, multimodal, and multilingual gaps identified above.

\section{Construction of MMTClinic}
The construction of \textbf\textit{{MMTClinic}} is outlined through the following key components: \textbf{dataset dimensions}, \textbf{detailed statistics}, and the \textbf{overall framework} that guided the creation of \textbf\textit{{MMTClinic}}. In Figure \ref{fig:mmt}, the overall construction of MMTClinic is presented in detail.


\subsection{Dataset Dimensions}
\textbf{\textit{MMTClinic}}  includes both multiple-choice (MCQ) and open-ended reasoning-based questions designed to evaluate clinical understanding from multimodal inputs. It is multilingual and covers five Indian languages, namely Hindi, Tamil, Marathi, English, and Bengali. It comprises three core tasks: \textbf{Task 1:} Given the physiological characteristics of a patient, such as heart rate, blood pressure, and respiratory rate, predict the in-hospital mortality outcome (0: survived, 1: died during hospital stay). \textbf{Task 2:} Given multivariate time-series data collected during the first 48 hours of ICU stay—including measurements like blood pressure, respiratory rate, and temperature-—forecast the patient’s heart rate (HR) as an early indicator of clinical deterioration. \textbf{Task 3:} Based on 48-hour physiological time-series data, including vital signs and lab measurements, estimate the Sequential Organ Failure Assessment (SOFA) score to quantify the level of organ dysfunction. Each task is evaluated under four distinct multimodal settings: (i) text and time-series (MCQ), (ii) text, image, and time-series (MCQ), (iii) text and time-series (reasoning), and (iv) text, image, and time-series (reasoning).

\begin{table*}[ht]
\centering
\begin{minipage}{0.48\textwidth}
\centering
\begin{tabular}{p{3cm}|c|c|c}
\hline
\textbf{Model} & \textbf{Task1} & \textbf{Task2} & \textbf{Task3} \\
\hline
\rowcolor{green!30} \multicolumn{4}{|l|}{\textbf{Small LLMs }} \\
\cellcolor{green!10}Mistral-7B-Instruct & 28.02 & 30.52 & 21.83 \\
\cellcolor{green!10}Qwen2.5-7B & 6.97 & 14.83 & 11.02 \\
\cellcolor{green!10}DeepSeek-R1-Llama-8B & 58.15 & 57.22 & 57.20 \\
\hline
\rowcolor{blue!25} \multicolumn{4}{|l|}{\textbf{Large LLMs }} \\
\cellcolor{blue!10}Llama-3.3-70B & 31.12 & 27.71 & 23.24 \\
\cellcolor{blue!10}Qwen3-30B-A3B & 43.77 & 42.92 & 31.84 \\
\cellcolor{blue!10}Qwen3-235B-A22B & 52.10 & 51.88 & 50.28 \\
\cellcolor{blue!10}QwQ-32B & 52.37 & 50.72 & 50.58 \\
\cellcolor{blue!10}DeepSeek-R1 & 66.0 & 65.1 & 65.3 \\
\hline
\rowcolor{yellow!40} \multicolumn{4}{|l|}{\textbf{Proprietary LLMs}} \\
\cellcolor{yellow!10}GPT-4.1-nano & 45.27 & 45.75 & 39.73 \\
\cellcolor{yellow!10}Gemini-2-flash & 27.4 & 35.53 & 30.47 \\
\hline
\end{tabular}
\caption{Task-wise average accuracy (MCQ, Text + Time Series)}
\label{tab:taskwise_mcq_avg}
\end{minipage}
\hfill
\begin{minipage}{0.48\textwidth}
\centering
\begin{tabular}{p{3cm}|c|c|c}
\hline
\textbf{Model} & \textbf{Task1} & \textbf{Task2} & \textbf{Task3} \\
\hline
\rowcolor{green!30} \multicolumn{4}{|l|}{\textbf{Small LLMs }} \\
\cellcolor{green!10}Mistral-7B-Instruct & 35.72 & 31.97 & 24.80 \\
\cellcolor{green!10}Qwen2.5-7B & 9.15 & 11.75 & 8.98 \\
\cellcolor{green!10}DeepSeek-R1-Llama-8B & 71.38 & 64.19 & 65.74 \\
\hline
\rowcolor{blue!25} \multicolumn{4}{|l|}{\textbf{Large LLMs }} \\
\cellcolor{blue!10}Llama-3.3-70B & 28.45 & 29.07 & 23.86 \\
\cellcolor{blue!10}Qwen3-30B-A3B & 35.99 & 37.02 & 26.06 \\
\cellcolor{blue!10}Qwen3-235B-A22B & 76.23 & 71.36 & 73.60 \\
\cellcolor{blue!10}QwQ-32B & 61.57 & 60.89 & 62.72 \\
\cellcolor{blue!10}DeepSeek-R1 & 64.61 & 61.25 & 64.35 \\
\hline
\rowcolor{yellow!40} \multicolumn{4}{|l|}{\textbf{Proprietary LLMs}} \\
\cellcolor{yellow!10}GPT-4.1-nano & 31.73 & 32.59 & 25.70 \\
\cellcolor{yellow!10}Gemini-2-flash & 31.04 & 32.34 & 26.15 \\
\hline
\end{tabular}
\caption{Task-wise average accuracy (Reasoning, Text + Time Series)}
\label{tab:taskwise_avg_accuracy}
\end{minipage}
\end{table*}

\begin{table*}[ht]
\centering
\begin{minipage}{0.48\textwidth}
\centering
\begin{tabular}{p{3cm}|c|c|c}
\hline
\textbf{Model} & \textbf{Task1} & \textbf{Task2} & \textbf{Task3} \\
\hline
\rowcolor{green!30} \multicolumn{4}{|l|}{\textbf{Small LLMs}} \\
\cellcolor{green!10}Qwen2.5-VL-7B & 30.13 & 33.28 & 25.83 \\
\hline
\rowcolor{blue!25} \multicolumn{4}{|l|}{\textbf{Large LLMs }} \\
\cellcolor{blue!10}LLaMA-3.2-11B-Vision-Instruct & 30.12 & 37.13 & 29.17 \\
\cellcolor{blue!10}Qwen2.5-VL-72B & 31.62 & 44.38 & 34.26 \\
\cellcolor{blue!10}Gemma-3-27B-IT & 26.91 & 29.68 & 42.91 \\
\hline
\rowcolor{yellow!40} \multicolumn{4}{|l|}{\textbf{Proprietary LLMs}} \\
\cellcolor{yellow!10}GPT-4.1-nano & 46.50 & 46.26 & 42.48 \\
\cellcolor{yellow!10}Gemini-2-Flash & 28.58 & 33.50 & 30.86 \\
\hline
\end{tabular}
\caption{Average MCQ Accuracy per Task (Text + Time Series + Image)}
\label{tab:mcq_image_grouped1}
\end{minipage}
\hfill
\begin{minipage}{0.48\textwidth}
\centering
\begin{tabular}{p{3cm}|c|c|c}
\hline
\textbf{Model} & \textbf{Task1} & \textbf{Task2} & \textbf{Task3} \\
\hline
\rowcolor{green!30} \multicolumn{4}{|l|}{\textbf{Small LLMs }} \\
\cellcolor{green!10}Qwen2.5-VL-7B & 33.07 & 38.76 & 42.39 \\
\hline
\rowcolor{blue!25} \multicolumn{4}{|l|}{\textbf{Large LLMs }} \\
\cellcolor{blue!10}LLaMA-3.2-11B-Vision-Instruct & 29.56 & 34.72 & 24.71 \\
\cellcolor{blue!10}Qwen2.5-VL-72B & 29.57 & 42.85 & 50.29 \\
\cellcolor{blue!10}Gemma-3-27B-IT & 22.30 & 25.36 & 45.15 \\
\hline
\rowcolor{yellow!40} \multicolumn{4}{|l|}{\textbf{Proprietary LLMs}} \\
\cellcolor{yellow!10}GPT-4.1-nano & 31.96 & 38.78 & 26.19 \\
\cellcolor{yellow!10}Gemini-2-Flash & 24.54 & 31.14 & 23.42 \\
\hline
\end{tabular}
\caption{Average Reasoning Accuracy per Task (Text + Time Series + Image)}
\label{tab:reasoning_image_grouped4}
\end{minipage}
\end{table*}

\subsection{Statistics of  MMTClinic}
The main statistics associated with MMTCLinic are shown below:\par
i) The MMTClinic benchmark comprises a total of 30,000 questions.\par
ii) Two distinct types of MCQ and open-ended reasoning-based questions.\par
iii) The data set spans four unique clinical tasks, each designed to evaluate different aspects of multimodal and multilingual clinical reasoning.\par
iv) MMTClinic supports 5 languages: English, Hindi, Bengali, Marathi, and Tamil.\par
v) Each task is associated with an average of 7,500 questions.\par
vi) On average, there are 6,000 questions available for each language. \par

\subsection{Framework for developing MMTClinic}
The following steps are used to develop \textbf{the MMTClinic} :\par

\begin{figure*}[!htbp]
\centering
  \includegraphics[scale=0.32]{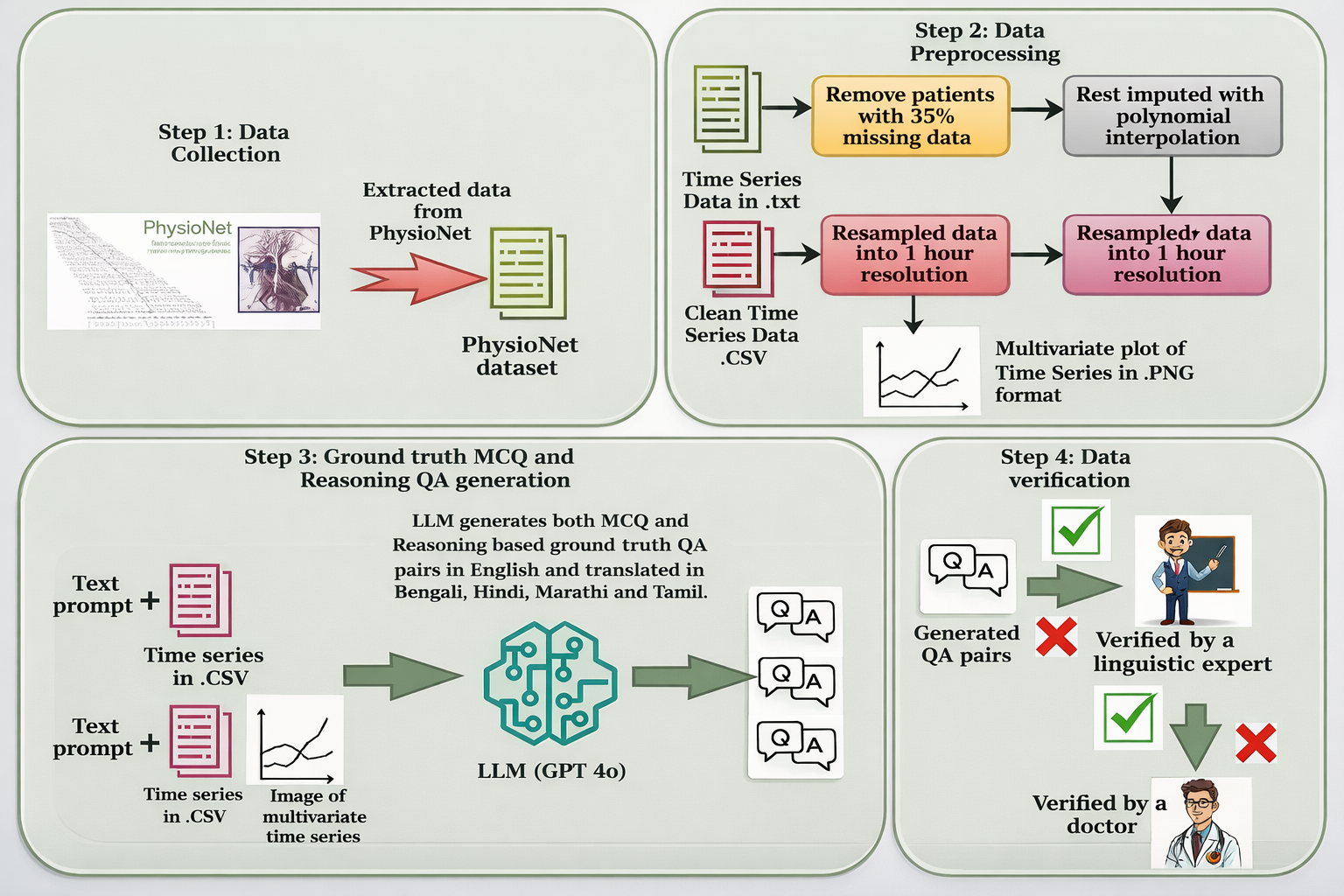}
  \caption{Construction of MMTClinic. Step 1 indicates data collection. Step 2 implies the detailed pre-processing steps for extracting complete data of a patient. Step 3 focused on question answer generation in two experimental setup (a) Text + Time Series and (b) Text + Time Series + Image. Step 4 indicates the question answer validation by medical and linguistic expert.   }
  \label{fig:mmt}
\end{figure*}

\textbf{Step 1: Data collection:}
We begin by obtaining multivariate clinical time series for ICU patients from the 2012 PhysioNet/Computing in Cardiology Challenge dataset, which provides up to 42 physiological and laboratory variables (e.g. HR, BP, RespRate, Temp, Glucose) recorded during the first 48 hours of stay of each patient. For \textbf{\textit{MMTClinic}} , we used hourly resampling and generated both numerical (.csv) and visual (.png) representations of six physiological signals. The scale, clinical richness, and structured signal data in this dataset make it an ideal choice for this benchmark.
\ref{fig:mmt}.

\textbf{Step 2: Data Pre-processing:}
To ensure consistency, patients with more than 35\% missing values are removed, and the remaining are imputed using multivariate polynomial interpolation. Signals originally sampled at irregular intervals are then uniformly resampled to 1-hour resolution.
Finally, both the cleaned numerical series (.csv) and the corresponding multivariate line-plot images (.png) are generated using the matplotlib library in Python. 

\textbf{Step 3: Initial Samples Generation:}
Our few-shot samples were carefully selected using a medical‑expert prompting strategy, ensuring at least one representative example from each task described in Appendix \ref{pmt}. Using these prompts, we generated 15,000 multiple-choice and 15,000 open-ended reasoning question-answer pairs in English from a randomly chosen subset of 2,000 patients. The Google Translate API was then used to translate these QA pairs into Bengali, Hindi, Marathi, and Tamil, after which professional linguists carefully reviewed and refined the translations to ensure accurate and high-quality outputs in all target languages.  \textbf{\textit{Importantly, the ground‑truth answers for both the MCQ and reasoning tasks were sourced directly from the PhysioNet dataset; no expert curation or guidance was used to derive these answers. Only task formation and samples question generation were done by experts.
}} The entire dataset including images, .CSV files, task wise QA pairs, and prompts can be accessed at \href{https://github.com/rakhujoy/MMTClinic}{this link}.

\textbf{Step 4: Samples Data Verification:}
To ensure the samples generated in Step 3 form a clinically relevant, high‑quality multilingual benchmark, we conduct a two‑stage evaluation: (1) clinical relevance review and (2) linguistic quality and fidelity assessment across languages. We discuss both of these aspects in detail in the following points:\par

\par

\textbf{Clinical Relevance Review:}  We used GPT‑4o to generate question–answer samples with few‑shot prompts curated by medical experts, and we took all ground‑truth answers directly from the source dataset. 
We conducted a manual expert review on a stratified random subset that included about 70\% of the samples from each task category. These categories were MCQ (Text + Time-Series), MCQ (Text + Time-Series + Images), Reasoning (Text + Time-Series), and Reasoning (Text + Images + Time-Series). Four medical experts independently reviewed each sample. The mean ratings were 4.2, 4.5, 4.4, and 4.5 for the respective tasks, indicating consistently high sample quality. This extensive study demonstrates that the carefully designed prompting strategy is effective in generating high-quality questions. Expert verification followed a standard set of guidelines to ensure consistency and objectivity. For each sample, reviewers evaluated (i) the clinical relevance of the question, (ii) the correctness of the ground-truth answer based on the underlying physiological time series, (iii) the clarity and plausibility of distractor options for multiple-choice questions, and (iv) the logical coherence of reasoning-based answers. Linguistic reviewers followed similar guidelines focusing on semantic accuracy, grammatical correctness, and the preservation of medical meaning across languages. Detailed verification instructions can be found in Appendix \ref{id}. Per task averages and inter‑annotator agreement are reported in the Appendix-\ref{qs}.

\textbf{Linguistic Quality:} 
To ensure high-quality multilingual translations,  the dataset was reviewed by native-speaking linguists who are fluent in the target languages. The linguistic evaluation followed a two-stage process:
\par
a) Each data point was rated on a scale from 0 to 5 based on linguistic accuracy and fluency. \par

b) Samples scoring below 3.5 were flagged for expert post-editing. Only a small portion required further refinement—approximately 3.8\% of samples in Tamil and Marathi, 2.1\% in Bengali, and 1\% in Hindi. \par

\begin{figure*}[t]
  \centering
  \begin{minipage}{0.4\linewidth}
    \centering
    \includegraphics[width=\linewidth]{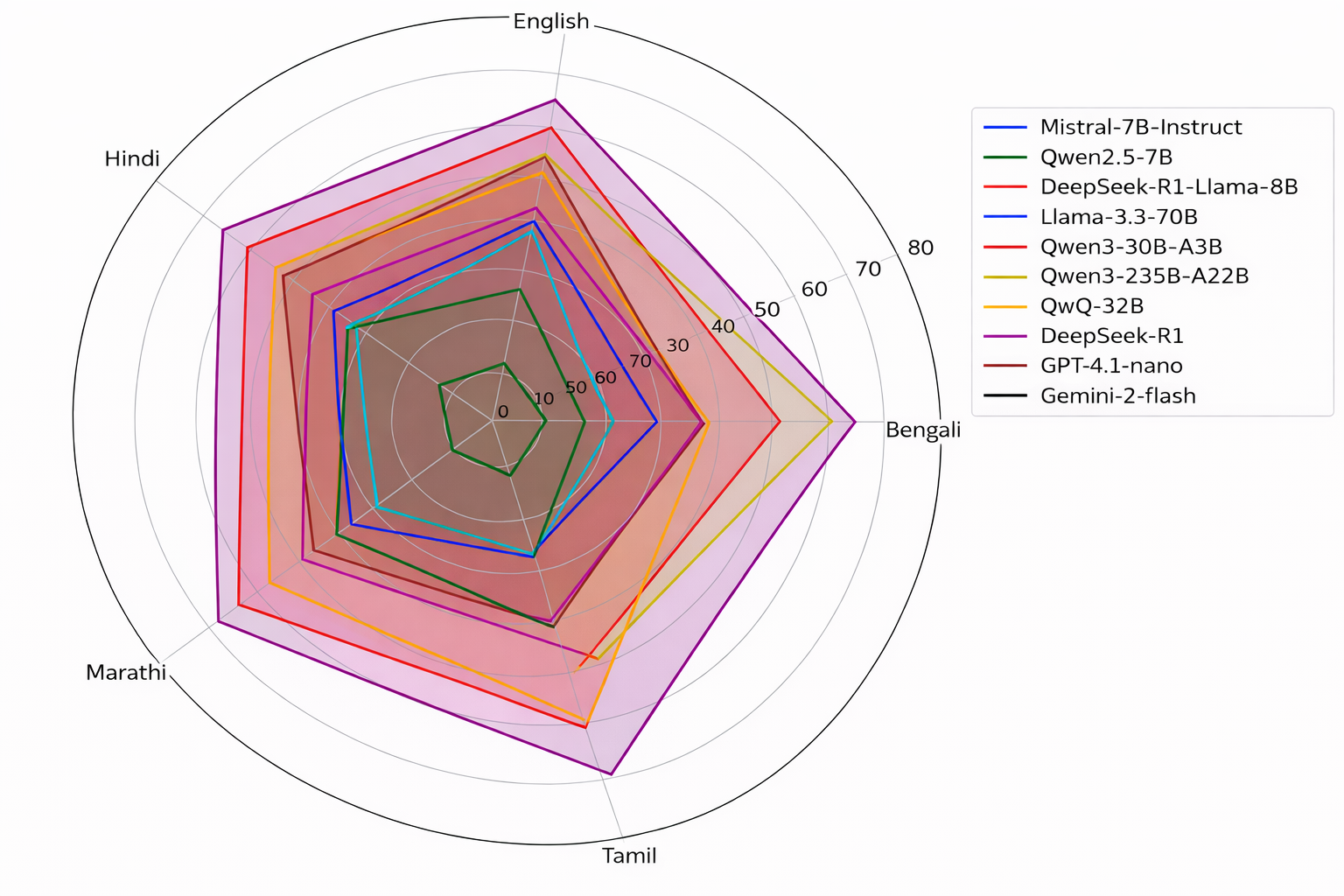}
    \caption*{(a)}
  \end{minipage}
  \hspace{0.02\linewidth} 
  \begin{minipage}{0.35\linewidth}
    \centering
    \includegraphics[width=\linewidth]{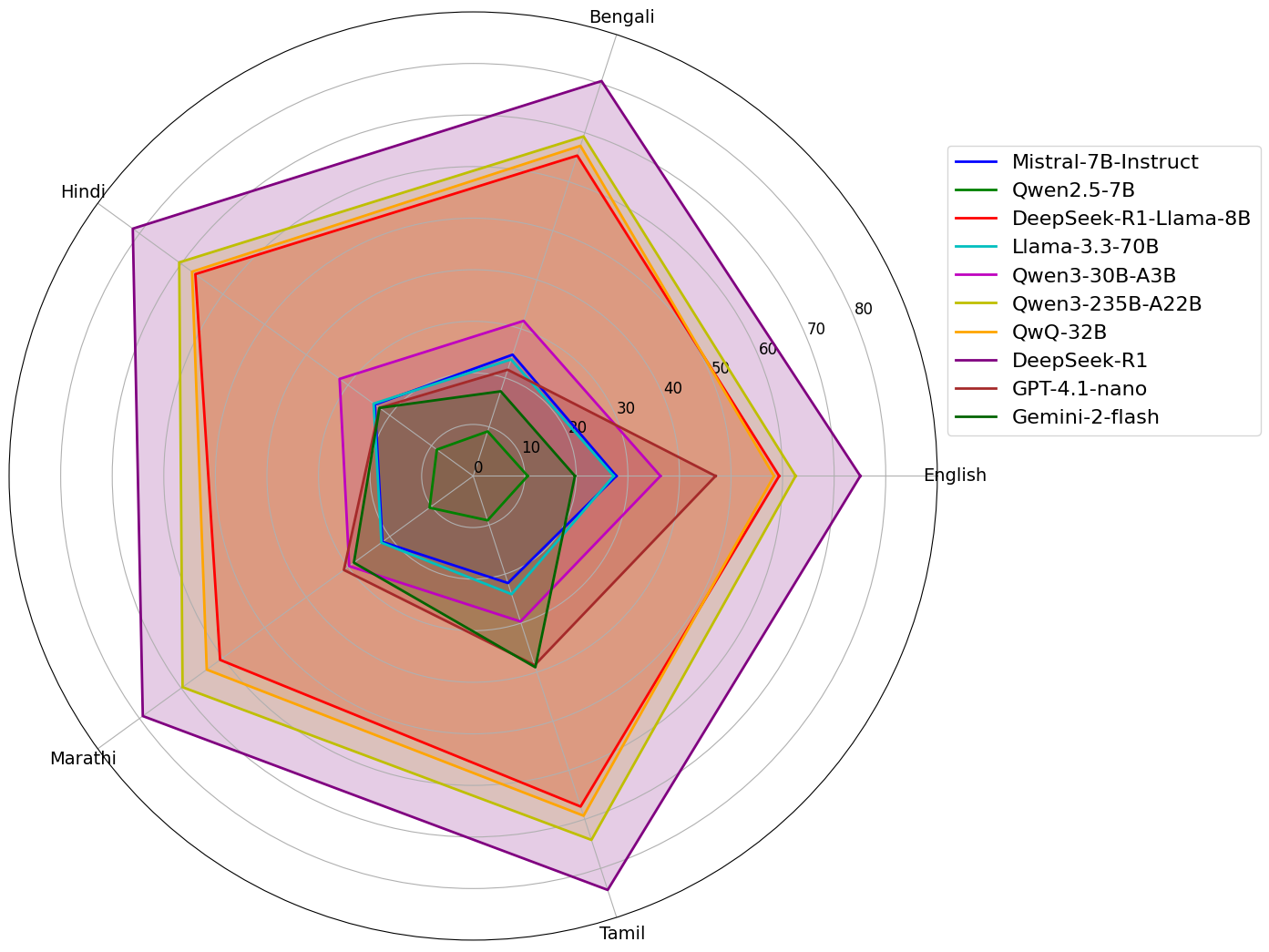}
    \caption*{(b)}
  \end{minipage}
  \caption{Radar plot indicating language-specific accuracy of LLMs for Text and Time Series modality: (a) MCQ and (b) Reasoning.}
  \label{fig:two_images1}
\end{figure*}

\begin{figure*}[t]
  \centering
  \begin{minipage}{0.35\linewidth}
    \centering
    \includegraphics[width=\linewidth]{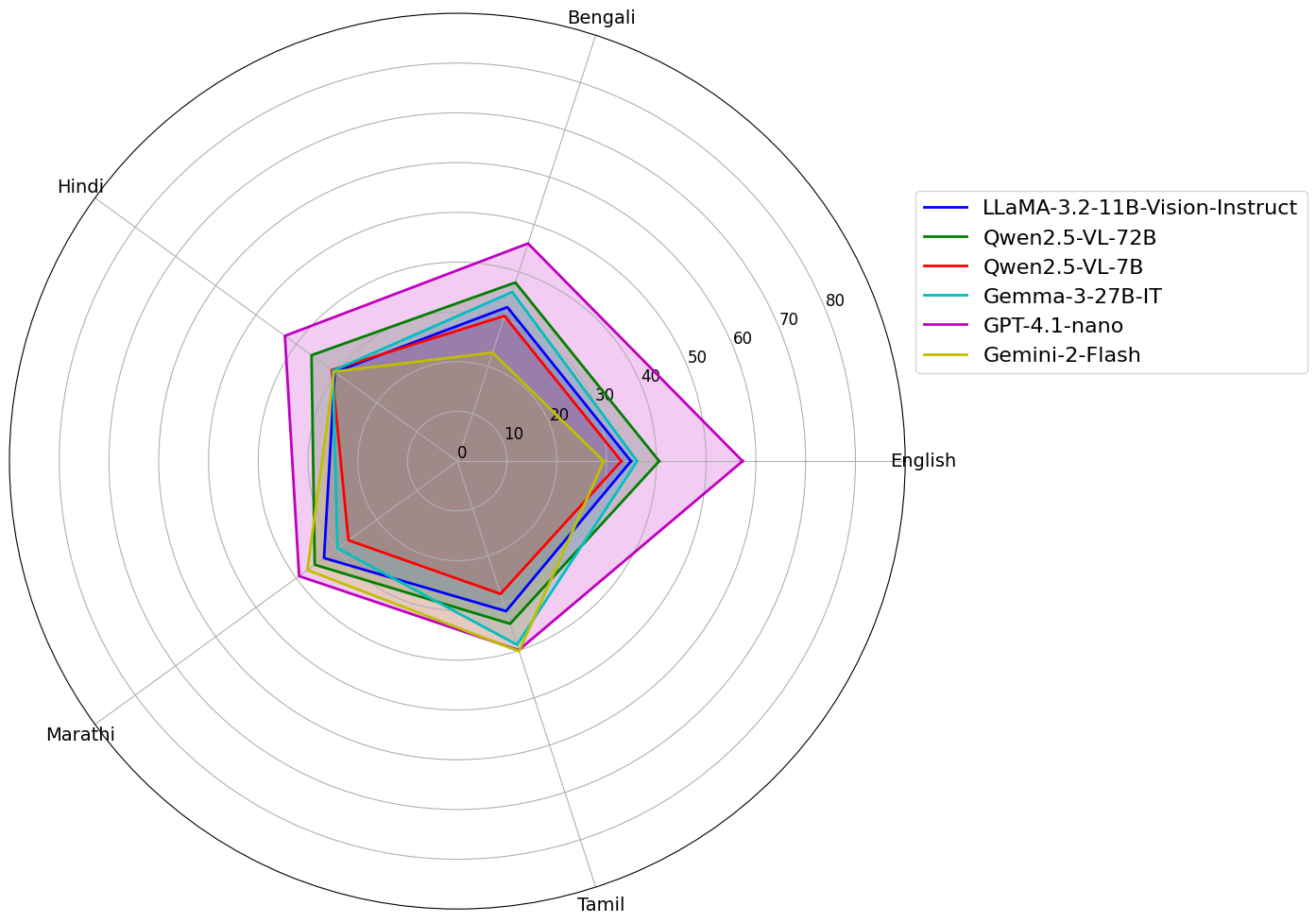}
    \caption*{(a)}
  \end{minipage}
  \hspace{0.02\linewidth}
  \begin{minipage}{0.35\linewidth}
    \centering
    \includegraphics[width=\linewidth]{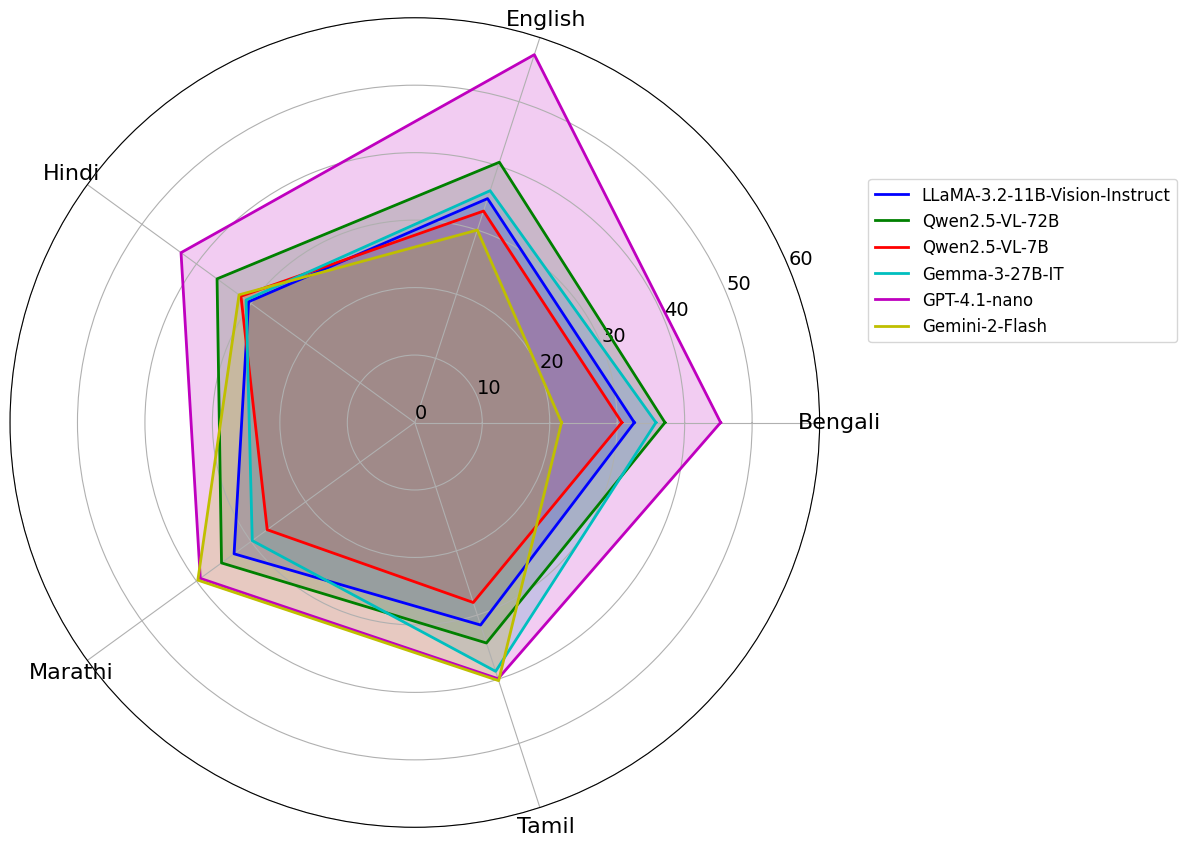}
    \caption*{(b)}
  \end{minipage}
  \caption{Radar plot indicating language-specific accuracy of LLMs for Text, Time Series, and Image modality: (a) MCQ and (b) Reasoning.}
  \label{fig:two_images}
\end{figure*}

\section{Experimental Section}

\subsection{Models}
To enable a comprehensive evaluation of diverse tasks in \textbf{MMTClinic}, which includes both \textbf{text + time-series} and \textbf{text + time-series + image} inputs, we benchmark a wide range of language models categorized into two primary groups: \textbf{text-only language models} and \textbf{multimodal language models}. Each group is further classified by scale (\textbf{small} vs. \textbf{large}) and accessibility (\textbf{proprietary} vs. \textbf{open}). For text-only models, we evaluate \textbf{Small Language Models (SLMs)} such as Mistral 7B Instruct \cite{samo2024fine}, Qwen 2.5 7B \cite{yang2024qwen2}, and DeepSeek R1-LLaMA 3 8B \cite{guo2025deepseek}, and \textbf{Large Language Models (LLMs)} including LLaMA 3.1 70B \cite{grattafiori2024llama}, Qwen 3 30B \cite{yang2025qwen3}, Qwen 2.5 35B \cite{hui2024qwen2}, QWQ 32B \cite{team2025qwq}, and DeepSeek R1-Large \cite{guo2025deepseek}. We also include proprietary models such as GPT-4.1 nano \cite{openai2023gpt4} and Gemini 2 Flash \cite{team2023gemini}. For the multimodal setting, we assess \textbf{Small Multimodal Models} like Qwen 2.5 VL 7B \cite{bai2025qwen2}, and \textbf{Large Multimodal Models} including LLaMA 3.2 11B Vision Instruct \cite{lee2025efficient}, Qwen 2.5 VL 72B \cite{bai2025qwen2}, and Gemma 3 27B \cite{team2023gemini}. The same proprietary models—GPT-4.1 nano \cite{openai_gpt41_nano_2025} and Gemini 2 Flash \cite{google_gemini2_flash_2025} are also considered in the multimodal evaluation.
The model parameters and hyper--parameters have been presented in Appendix \ref{mp} and \ref{sec:appendix21}.

\subsection{Evaluation Setup}
We have conducted \textbf{zero-shot}, \textbf{few-shot}, and \textbf{CoT} based evaluation across all four sub tasks defined in \textbf{MMTClinic} to authentically assess the LLM's inherent medical knowledge and reasoning capabilities. This approach offers a more reliable and unbiased evaluation of the LLMs ability to generalize to unseen clinical questions. To comprehensively assess model performance, we divide the analysis into four distinct settings: (i) \textbf{MCQ with Text + Time-Series}, (ii) \textbf{Reasoning with Text + Time-Series}, (iii) \textbf{MCQ with Text + Time-Series + Image}, and (iv) \textbf{Reasoning with Text + Time-Series + Image}. In addition to these modality-based settings, we perform a fine-grained evaluation across all four types of question formats, further disaggregated by individual language, to examine multilingual generalization capabilities.

\subsection{Discussion on Results}

\textbf{Performance across different settings:} Across the four evaluation settings, we see a clear hierarchy of model strengths and weaknesses as discussed in points below. \textbf{a) MCQ + Text + Time-Series:} From Table \ref{tab:taskwise_mcq_avg},  Small LLMs like DeepSeek-R1 and LLaMA-8B show better performance, with average accuracies above 57\% compared to Qwen 2.5-7B, which is around 7\%, and Mistral 7B, which is about 28\%. Among the larger models, Qwen 3-235B and QWQ-32B excel, achieving robust accuracy in the mid-60\% range. They outperform proprietary models like GPT-4.1-Nano, which is roughly 45\%, and Gemini 2-Flash by 27\%. These proprietary models fall between the capabilities of small and large open-source LLMs. 

\begin{figure*}[t]
  \centering
  \begin{minipage}{0.45\linewidth}
    \centering
    \includegraphics[width=\linewidth]{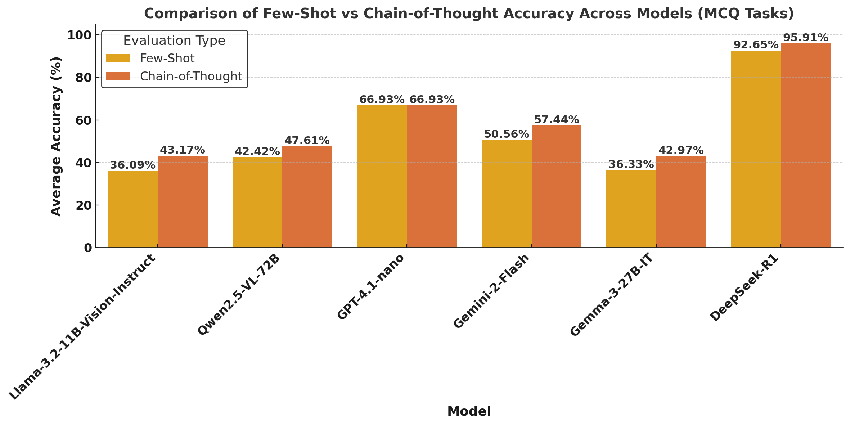}
    \caption*{(a)}
  \end{minipage}
  \hfill
  \begin{minipage}{0.45\linewidth}
    \centering
    \includegraphics[width=\linewidth]{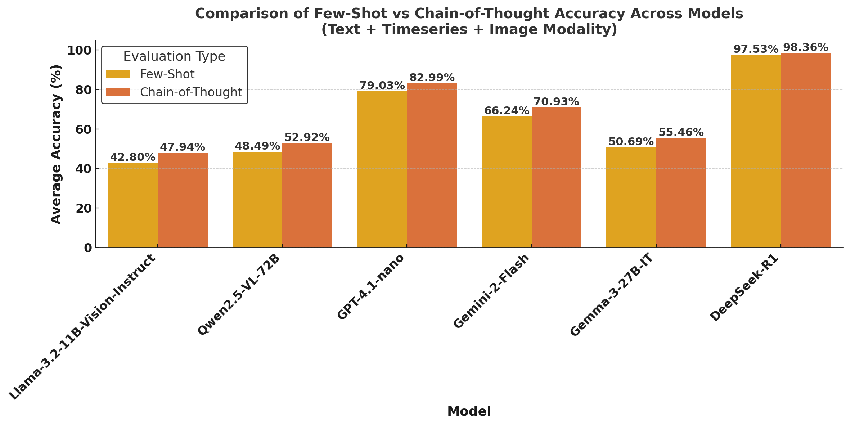}
    \caption*{(b)}
  \end{minipage}
  \caption{Box plot indicating few-shot and CoT specific accuracy of LLMs for Text and Time Series modality (a) MCQ (b) Reasoning }
  \label{fig:two_images100}
\end{figure*}
\begin{figure}
  \centering
  \includegraphics[scale=0.62]{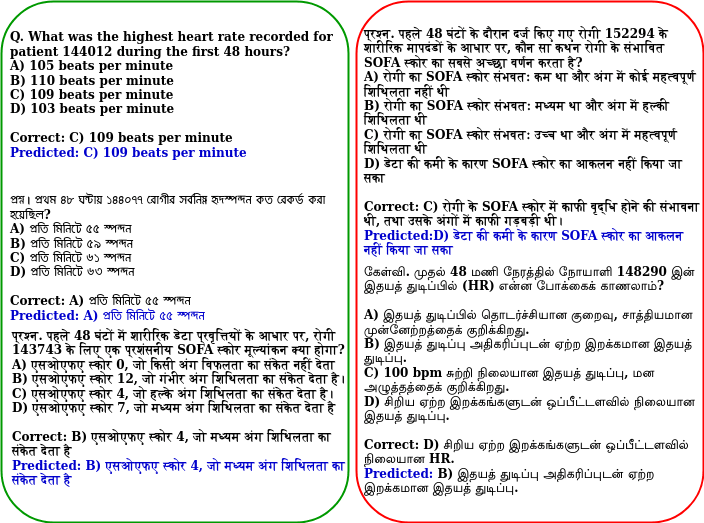}
  \caption{The LHS displays the correctly answered questions, while the RHS highlights the incorrectly answered ones by the language models on the MMTClinic dataset.}
  \label{fig:error}
\end{figure}

\textbf{b) Reasoning + Text + Time-Series:} 
In Table \ref{tab:taskwise_avg_accuracy}, the average accuracy for reasoning tasks using Text and Time-Series shows a growing performance gap. DeepSeek-R1-LLaMA-8B achieves about 71\%, while Qwen 3-235B leads at around 76\%. In contrast, Qwen 2.5-7B stays below 10\%, and proprietary models fall to the low-30\% range. This highlights how open-source reasoning LLMs better manage temporal clinical data.

\textbf{c) MCQ + Text + Time-Series + Image:} Table \ref{tab:mcq_image_grouped1} shows that adding images results in a notable performance drop. GPT-4.1-nano performs best at roughly 46\%, followed by Qwen 2.5-VL-7B and 72B at 30--32\%. Other vision-language models, including LLaMA-3.2 Vision and Gemma~3~27B, score lower at around 26--30\%. This suggests that integrating visual elements remains difficult and proprietary models handle it better.

\textbf{d) Reasoning + Text + Time-Series + Image:} This is the most challenging setting (Table \ref{tab:reasoning_image_grouped4}). Qwen 2.5-VL~7B slightly outperforms GPT-4.1-nano (33\% vs.\ 32\%), while larger multimodal models and Gemini~2-Flash fall into the low-to-mid-20\% range. Overall, models with specialized multimodal training do better on image-based MCQs, while temporal reasoning models such as DeepSeek variants and Qwen~3-235B excel in text-time tasks. Simpler models, such as Qwen~2.5-7B and Mistral~7B, consistently perform poorly.

\par

\begin{tcolorbox}[colback=green!10!white, colframe=green!50!black, boxrule=0.5pt, arc=2pt, left=4pt, right=4pt, top=4pt, bottom=4pt]
\small
\textbf{\textit{The key conclusion from the task-wise analysis is that open-source LLMs outperform in text-based MCQs and reasoning tasks, whereas proprietary models demonstrate superior performance on image-based MCQs and multimodal reasoning.}}
\end{tcolorbox}

\textbf{Performance of models across languages:} For the \textbf{MCQ (Text + Time-Series)} task in Figure \ref{fig:two_images1} (a), English leads, followed closely by Bengali and Marathi. Large models like Qwen 3-235B and DeepSeek-R1-LLaMA-8B show strong multilingual consistency, losing only 5 to 10\% in regional languages. In contrast, smaller models, such as Mistral 7B and Qwen 2.5-7B, drop over 30\% in Tamil and Hindi. GPT-4.1-nano maintains stability across all five languages.  In the \textbf{Reasoning (Text + Time-Series)} task (Figure \ref{fig:two_images1} (b)), the overall precision decreases and the disparities grow. Top models still score at least 70\% in English and Bengali, but they fall 15 to 20\% in Tamil and Marathi. Mid-sized models experience sharper declines, while proprietary models remain stable at 30 to 35\%. 
In the \textbf{MCQ (Text + Time-Series + Image)} setting (Figure \ref{fig:two_images} (a)), GPT-4.1-nano leads with scores of 45 to 50\% in English, showing a moderate 10 to 12\% drop in regional performance. Qwen 2.5-VL-7B performs moderately, scoring 30 to 35\%, while other open-source vision-instruct models cluster between 25 to 30\%. Simpler models fall below 20\%.  For \textbf{Reasoning (Text + Time-Series + Image)} in Figure \ref{fig:two_images} (b), accuracy drops another 5 to 8\%. GPT-4.1-nano stays at 35 to 40\%, while Qwen 2.5-VL-7B drops from about 38\% in English to 20\% in other languages. Gemini 2-Flash and vision-instruct models stabilize around 24 to 32\%, reflecting limited but consistent multimodal reasoning.\par

\begin{tcolorbox}[colback=green!10!white, colframe=green!50!black, boxrule=0.5pt, arc=2pt, left=4pt, right=4pt, top=4pt, bottom=4pt]
\small
\textbf{\textit{The key conclusion is that proprietary models are stronger in image-related tasks compared to open weight models. Models across all tiers experienced the worst performance in reasoning-based tasks, especially in languages like Tamil.}}
\end{tcolorbox}

\textbf{Performance across model types:}
Small language models consistently perform poorly, rarely exceeding 30\% in MCQ and falling below 10\% in reasoning due to their limited capacity and lack of multimodal pretraining. LLMs do well on text and time-series tasks, with MCQ accuracy between 50\% and 70\% and reasoning up to 75\%. However, they drop 10 to 20\% when images are added, showing their limited ability to generalize to multimodal inputs. Proprietary Models (e.g., GPT-4.1-nano, Gemini 2-Flash) provide stable multilingual performance with narrow variance (about 10 to 12\%) and lead in multimodal MCQs (about 45\%) and reasoning (30 to 40\%), although the top open-source LLMs outperform them in text-only settings. Multimodal Models (vision-instruct and VL variants) manage visual inputs the best, reaching 30 to 35\% MCQ and 25 to 32\% reasoning accuracy with images. They are the only group to outperform proprietary models in multimodal evaluations, even though they lag behind in pure time-series tasks.

\begin{tcolorbox}[colback=green!10!white, colframe=green!50!black, boxrule=0.5pt, arc=2pt, left=4pt, right=4pt, top=4pt, bottom=4pt]
\small
\textbf{\textit{Multimodal models outperform others on image-based tasks, while proprietary LLMs lead in multilingual consistency and open-source LLMs excel in text-only reasoning.}}
\end{tcolorbox}

\textbf{Performance across prompting techniques:}
Figures \ref{fig:two_images100}(a) and \ref{fig:two_images100}(b) compare Few-Shot and Chain-of-Thought (CoT) prompting across clinical tasks with increasing modality complexity—(a) text + time-series and (b) text + time-series + vision. DeepSeek-R1 consistently performs best, reaching up to 98.36\% accuracy with negligible CoT gains, indicating strong inherent reasoning. GPT-4.1-nano shows limited improvement in MCQs but benefits modestly from CoT in multimodal settings (+3.96\%). Gemini-2-Flash exhibits the largest CoT-driven gains in MCQs (+6.88\%), suggesting high responsiveness to structured reasoning. In contrast, Gemma-3-27B-IT shows minimal adaptation. Vision-oriented models such as LLaMA-3.2-11B-Vision-Instruct and Qwen2.5-VL-72B perform poorly with negligible CoT benefits, highlighting shallow reasoning and weak generalization. Overall, the results indicate that only a small subset of models can effectively exploit CoT prompting across complex clinical modalities.\par

\begin{tcolorbox}[colback=green!10!white, colframe=green!50!black, boxrule=0.5pt, arc=2pt, left=4pt, right=4pt, top=4pt, bottom=4pt]
\small
\textbf{\textit{DeepSeek-R1 consistently outperforms all models across both multimodal and multilingual clinical tasks with minimal reliance on Chain-of-Thought prompting, while others—especially proprietary and vision-centric models—exhibit limited generalization and rely heavily on CoT to boost reasoning.}}
\end{tcolorbox}
Extended experimental results and performance comparison of the MMTClinic compared to state-of-the-art benchmarks are shown in the Appendices \ref{rs} and \ref{sec:appendix2}.

\section{Error Analysis}
To better understand the strengths and limitations of the best performing LLMs in the MMTClinic data set, we conducted an error analysis by categorizing the questions into correctly and incorrectly answered groups, as illustrated in Figure \ref{fig:error}. The analysis reveals that questions on the left side tend to be well-structured and clearly framed, whereas those on the right suffer from issues such as limited subjectivity, insufficient contextual grounding in data, and semantic overlap. 
These shortcomings indicates that incorrectly answered questions typically require deeper temporal abstraction, longer-range trend integration, or joint reasoning across modalities. These errors highlight current model limitations rather than deficiencies in question construction, as all samples were clinically validated validated and based on original data.

\section{Conclusion:}  

We present MMTClinic, a clinically validated benchmark for evaluating multimodal and multilingual reasoning in large language models across text, physiological time-series, and visual data in five languages. Experiments on 13 state-of-the-art models show that while some open-source LLMs handle time-series reasoning well, proprietary models excel in complex multimodal settings. However, consistent performance drops in vision-language reasoning and regional languages reveal persistent challenges in multimodal integration and cross-lingual generalization. These gaps reflect genuine reasoning limitations rather than dataset artifacts, making MMTClinic a reliable foundation for advancing clinically grounded multimodal AI.

\section{Limitations}
The limitations of this work are stated below :\par
\textbf{Limited Language Coverage:} \textbf{\textit{MMTClinic}} includes only five Indian languages (English, Hindi, Bengali, Marathi, Tamil), which may not generalize to other global or low-resource languages. \par

\textbf{Single Data Source:} The benchmark relies solely on the 2012 PhysioNet/CinC ICU dataset, limiting evaluation to one institution’s patient population and measurement protocols .\par

\textbf{Static Visual Modality:} Visual inputs are restricted to line‐plot images of time‐series trends, excluding richer imaging modalities (e.g., radiographs, CT scans) commonly encountered in clinical practice .\par


\textbf{Automatic Translation Artifacts:} While translations into regional languages were reviewed by experts ‐, reliance on machine translation (Google Translate) may introduce certain subtle errors or biases in the phrasing of the questions that affect model performance. \par

\textbf{Accuracy Only Metrics:} Evaluation uses accuracy for both MCQ and reasoning tasks, which does not capture other aspects of model performance such as calibration, reasoning depth, or clinical safety. \par

\textbf{Discrete Answer Formats:} The current benchmark evaluates reasoning mainly through discrete outputs, such as multiple-choice questions and binary responses. While this approach allows for scalable and objective comparisons, it does not directly measure the quality of explanations or their interpretability. We see this as a trade-off between reliable evaluation and transparent reasoning. Incorporating free-form rationales and chain-of-thought supervision remains an important area for future work.


\section{Ethical Considerations}
Use of Proprietary Models and Data Compliance: Each proprietary model was used carefully in accordance with the access guidelines that each provider had provided. Crucially, no personally identifiable or protected health information was accessed or shared at any point during the study's use of the PhysioNet 2012 dataset, which is completely de-identified and openly accessible. In order to fully adhere to HIPAA compliance and data privacy standards, evaluations were conducted exclusively on derived QA content using synthetic prompts.

\section{Appendix}
\label{app}
The Appendix provides additional details on Frequently Asked Questions \ref{sec:Freq_Ques}, model parameters \ref{mp}, model hyper--parameters \ref{sec:appendix21}, performance comparison compared to benchmarking datasets \ref{sec:appendix2}, performance of different prompting techniques across languages \ref{pd}, questions \ref{qs}, instructions for the doctor for Validation \ref{id}, results \ref{rs}, and prompts \ref{pmt}.
\par

\subsection{Frequently Asked Questions (FAQ)}
\label{sec:Freq_Ques}

\textbf{Q1: How clinically realistic are the questions if they are generated using LLM prompting?} \par
\textbf{Ans:} Although the questions are generated using an LLM-based prompting pipeline, all ground-truth answers are directly sourced from the PhysioNet clinical dataset, and both MCQ and reasoning questions are curated to reflect realistic ICU scenarios. Additionally, a stratified subset of samples was reviewed by medical experts, who assigned high average clinical relevance scores, indicating that the generated questions are clinically meaningful rather than templated artifacts. \par

\textbf{Q2: Are the ground-truth answers independent of the LLM generation process?} \par
\textbf{Ans:} Yes. The LLM is used only to generate the question formulations, while all ground-truth answers are derived directly from the original physiological time-series and outcome files in the PhysioNet dataset. No expert curation or model-generated reasoning is used to derive answers, ensuring independence between question generation and answer supervision. \par

\textbf{Q3: Does reliance on a single ICU dataset limit generalizability?} \par
\textbf{Ans:} While MMTClinic is constructed using the PhysioNet 2012 ICU dataset, this choice enables controlled benchmarking with well-validated clinical signals. The benchmark is intended as an evaluation resource rather than a deployment dataset, and we view single-source construction as a common and acceptable trade-off for reproducibility and standardization in clinical benchmarks. \par

\textbf{Q4: What additional value do line-plot images provide beyond numeric time-series?} \par
\textbf{Ans:} The visual modality explicitly tests whether models can jointly reason over numerical trends and their visual abstractions. The observed performance drop when images are introduced highlights current limitations in multimodal temporal reasoning rather than redundancy, making this a deliberate stress test for multimodal models. \par

\textbf{Q5: Why are richer clinical imaging modalities not included?} \par
\textbf{Ans:} MMTClinic focuses on early ICU monitoring where physiological trends play a dominant role. Line-plot visualizations provide a controlled visual modality tightly aligned with time-series signals, allowing focused evaluation of temporal reasoning without confounding factors introduced by unrelated imaging tasks. \par

\textbf{Q6: Why were only five Indian languages selected?} \par
\textbf{Ans:} The selected languages represent linguistically diverse and underexplored Indic languages with substantial real-world clinical relevance. Our goal is not exhaustive multilingual coverage but to provide a fair and challenging testbed for low-resource India-specific multilingual clinical reasoning, which remains largely absent in existing benchmarks. \par

\textbf{Q7: Could machine translation artifacts affect multilingual evaluation?} \par
\textbf{Ans:} While initial translations were generated using Google Translate, all non-English samples were reviewed and refined by native-speaking linguists. Only a small fraction required post-editing, and expert ratings indicate consistently high linguistic quality across languages. \par

\textbf{Q8: Why is accuracy the primary evaluation metric for reasoning tasks?} \par
\textbf{A:} Accuracy enables objective and scalable comparison across models, languages, and modalities. Given the clinical domain, correctness of the final decision is critical, and accuracy provides a conservative baseline metric. We acknowledge that explanation quality is important and discuss it as future work. \par

\textbf{Q9: Why are free-form rationales not explicitly evaluated?} \par
\textbf{A:} Evaluating free-form explanations introduces subjectivity and scalability challenges. MMTClinic prioritizes reliable benchmarking of decision correctness. Incorporating rationale, faithfulness, and explanation quality is an important direction we leave for future extensions. \par

\textbf{Q10: How robust are the reported results statistically?} \par
\textbf{A:} All models are evaluated under identical inference settings across tasks and languages. While we report single-run inference results following standard benchmarking practice, the large evaluation scale (30K samples) mitigates variance and provides stable comparative trends. \par

\textbf{Q11: Are comparisons between open-source and proprietary models fair?} \par
\textbf{A:} We evaluate all models in inference-only mode using publicly available checkpoints or provider APIs with default settings. The goal is not to equalize training conditions, but to assess real-world performance differences under practical usage scenarios. \par

\textbf{Q12: Does the benchmark favor reasoning-optimized models?} \par
\textbf{A:} MMTClinic intentionally includes tasks that require temporal abstraction and cross-modal integration. Strong performance by reasoning-optimized models highlights genuine capability differences rather than benchmark bias, aligning with the benchmark’s stated objectives. \par

\textbf{Q13: Are models performing true reasoning or exploiting dataset patterns?} \par
\textbf{A:} Error analysis shows that failures predominantly occur on questions requiring long-range temporal integration and multimodal grounding, suggesting that models are not simply exploiting surface correlations. All questions are grounded in original clinical data rather than synthetic labels. \par

\textbf{Q14: How should MMTClinic be used by future researchers?} \par
\textbf{A:} MMTClinic is designed as an evaluation benchmark to diagnose multimodal, India specific, and time-series reasoning gaps. It should not be interpreted as a clinical decision-support system but as a controlled testbed for advancing clinically grounded AI research. \par

\textbf{Q15: What safeguards prevent misuse of the benchmark for unsafe clinical automation?} \par
\textbf{A:} The dataset is fully de-identified, and all evaluations are conducted on synthetic QA formulations derived from real data. We explicitly state that MMTClinic is not intended for deployment or automated decision-making without human oversight. \par

\subsection{Model parameters}
\label{mp}
Model parameters has been enlisted in Table \ref{tab:llm_parameters_only}.
\begin{table}[ht]
  \centering
  \begin{tabular}{p{5.5cm}|c}
    \hline
    \textbf{LLMs} & \textbf{Parameters} \\
    \hline
    DeepSeek-R1 \cite{guo2025deepseek} & Commercial \\
     GPT-4.1-nano \cite{popov2025transferring} & Commercial \\
     gemini-2-flash\cite{rakshith2025aicoe}& Commercial \\
    DeepSeek-R1-Llama-8B \cite{guo2025temporal}& 8B \\
   
    Gemma-3-27B-IT \cite{team2025gemma}& 27B \\
    Llama-3.2-11B-Vision-Instruct \cite{lee2025efficient} & 11B \\
    Llama-3.3-70B \cite{wihl2025data} & 70B \\
    Mistral-7B-Instruct \cite{samo2024fine}& 7B \\
    
    QwQ-32B \cite{zheng2024processbench}& 32B \\
    Qwen2.5-7B \cite{yang2024qwen2} & 7B \\
    Qwen2.5-VL-7B \cite{bai2025qwen2}& 7B \\
    Qwen3-235B-A22B \cite{ji2025thinking}& 235B \\
    Qwen3-30B-A3B\cite{yang2025qwen3} & 30B \\
    \hline
  \end{tabular}
  \caption{List of LLMs with corresponding parameter sizes}
  \label{tab:llm_parameters_only}
\end{table}

\subsection{Inference Setup and Hyperparameters}
\label{sec:appendix21}
All evaluated models operated in inference-only mode with publicly available or provider-hosted checkpoints. We did not perform any additional fine-tuning. For open-source models, we used PyTorch with HuggingFace Transformers and vLLM backends on NVIDIA A100 GPUs (80 GB VRAM) for inference. Unless stated otherwise, evaluations took place on a single GPU for each model instance. We used greedy decoding for all open-source models with a temperature of 0.0, top-p of 1.0, and a maximum output length of 512 tokens to guarantee consistent evaluation. We evaluated commercial models (like GPT-4o, Gemini, Claude) using their APIs with default inference settings.

\subsection{Performance comparison compared to the state-of-the-art benchmarking datasets}
\label{sec:appendix2}

In Table \ref{tab:benchmark_comparison}, the proposed dataset is compared against the clinical QA and reasoning benchmarks across key dimensions.
\begin{table*}[t]
\centering
\scriptsize
\setlength{\tabcolsep}{2.5pt}
\renewcommand{\arraystretch}{1.1}

\caption{Comparison of clinical QA and reasoning benchmarks across key dimensions.}
\label{tab:benchmark_comparison}

\begin{tabular}{|l|c|c|c|c|c|c|c|c|}
\hline
\textbf{Dataset} & \textbf{\#Samples} & \textbf{Modalities} &
\textbf{Question Types} & \textbf{Multi.} & \textbf{Reason.} &
\textbf{MCQ} & \textbf{\#Tasks} & \textbf{\#Langs.} \\
\hline
ECG-QA & $\sim$2K & ECG Signal & MCQ & No & No & Yes & 1 & 1 \\
\hline
ECG-Expert-QA & 47K & ECG Signal & MCQ + Open-ended & No & Yes & Yes & Multi-cond. & 1 \\
\hline
Q-Heart & 8.7K & ECG + Text & Open-ended & No & Yes & No & Multi-cond. & 1 \\
\hline
TIMER-Bench & $\sim$10K & Text + TS & Instructional QA & No & Yes & No & Temporal QA & 1 \\
\hline
EHRNoteQA & $\sim$2K & Text + TS & Open-ended & No & Yes & No & Patient QA & 1 \\
\hline
EHRXQA & N/S & Text + Image + TS & Open-ended & No & Yes & No & EHR + X-ray & 1 \\
\hline
MedFuse Benchmark & $\sim$35K & TS + Image & Classification & No & No & No & 3 & 1 \\
\hline
\textbf{MMTClinic} & \textbf{30K} & \textbf{Text + TS + Image} &
\textbf{MCQ + Reasoning} & \textbf{Yes} & \textbf{Yes} &
\textbf{Yes} & \textbf{3} & \textbf{5} \\
\hline
\end{tabular}
\end{table*}

\subsection{Performance of different prompting techniques across languages:}
\label{pd}
\begin{figure*}[!htbp]
  \centering
  \begin{minipage}{0.90\linewidth}
    \centering
    \includegraphics[width=\linewidth]{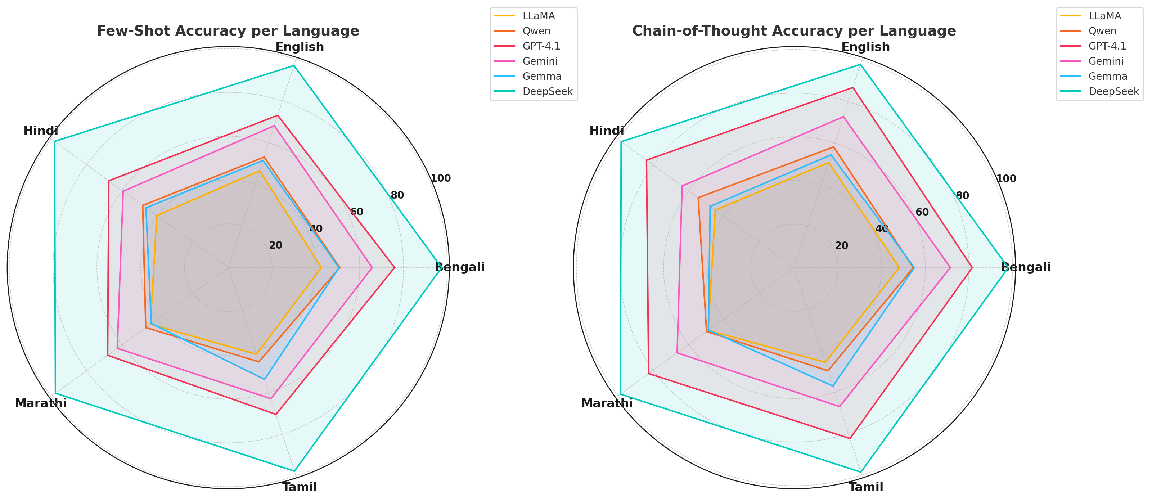}
    \caption*{(a)}
  \end{minipage}
  \hfill
  \begin{minipage}{0.90\linewidth}
    \centering
    \includegraphics[width=\linewidth]{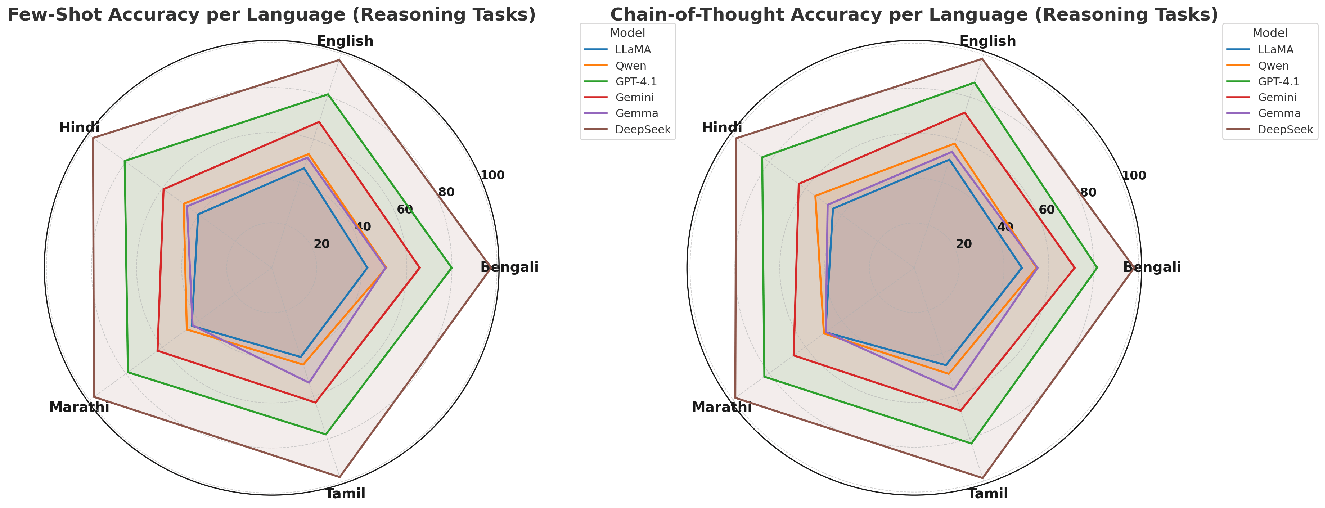}
    \caption*{(b)}
  \end{minipage}
  \caption{Radial plot indicating language-specific accuracy of LLMs under few-shot and CoT setup for Text, Time Series, and Image modality (a) MCQ (b) Reasoning }
  \label{fig:two_images101}
\end{figure*}

Figures \ref{fig:two_images101} (a) and (b) present a radar-based comparison of LLMs on multilingual reasoning tasks in five languages. They showcase model strength, cross-lingual generalization, and the effects of Few-Shot versus Chain-of-Thought prompting. Observations indicate that DeepSeek-R1 leads in all languages and setups, outperforming others by 8 to 20\%. It displays consistent accuracy with a small CoT gain of about 1 to 2\%. GPT-4.1-nano follows, lagging by 10 to 12\%, but shows better Chain-of-Thought responsiveness, especially in Hindi and Bengali, where it gains 6 to 8\%. This suggests a dependence on structured reasoning. Gemini-2-Flash exceeds lower-tier models by 5 to 10\%, achieving Chain-of-Thought gains of 6 to 9\% and showing moderate multilingual ability. Qwen2.5-VL-72B and LLaMA-3.2-11B-Vision-Instruct fall behind by 15 to 25\%, with limited Chain-of-Thought benefits of about 2 to 4\%. This points to weak multilingual reasoning. Gemma-3-27B-IT performs the worst, with a gap of 30 to 35\% in Tamil and Marathi. This reveals a significant bias in language and reasoning. Overall, DeepSeek-R1 excels in language variety and reasoning depth, while others rely more on CoT to compensate for foundational weaknesses.

\subsection{Questions}
\label{qs}
Table \ref{tab:expert_ratings_clean} presents detailed per-task evaluations from medical and linguistic experts, reporting high consistency across all QA types. Mean ratings ranged from 4.2 to 4.5, indicating strong clinical relevance and linguistic quality of the generated samples.
\begin{table}[!htbp]
\centering
\small
\begin{tabular}{|p{6cm}|}
\hline
\rowcolor{gray!20}
\textbf{Per-Task Expert Ratings (Medical, Linguistic, Mean)} \\
\hline
\textbf{MCQ (Text + TimeSeries)} \\
\hspace{1em}• Medical Expert Rating: \textbf{4.3} \\
\hspace{1em}• Linguistic Expert Rating: \textbf{4.1} \\
\hspace{1em}• Mean Rating: \textbf{4.2} \\
\hline
\textbf{MCQ (Text + TimeSeries + Images)} \\
\hspace{1em}• Medical Expert Rating: \textbf{4.6} \\
\hspace{1em}• Linguistic Expert Rating: \textbf{4.4} \\
\hspace{1em}• Mean Rating: \textbf{4.5} \\
\hline
\textbf{Reasoning (Text + TimeSeries)} \\
\hspace{1em}• Medical Expert Rating: \textbf{4.5} \\
\hspace{1em}• Linguistic Expert Rating: \textbf{4.3} \\
\hspace{1em}• Mean Rating: \textbf{4.4} \\
\hline
\textbf{Reasoning (Text + TimeSeries + Images)} \\
\hspace{1em}• Medical Expert Rating: \textbf{4.6} \\
\hspace{1em}• Linguistic Expert Rating: \textbf{4.4} \\
\hspace{1em}• Mean Rating: \textbf{4.5} \\
\hline
\end{tabular}
\caption{Vertically formatted expert review scores for each task.}
\label{tab:expert_ratings_clean}
\end{table}
\subsection{Instructions for the Doctor for Validation}
\label{id}

The clinical validation of MMTClinic was done by four medical professionals with significant experience in critical care medicine. They hold postgraduate medical degrees and have 8 to 12 years of clinical experience in intensive care units. They currently work as attending physicians at IIT Patna. Their daily responsibilities include monitoring patients in the ICU, and interpreting physiological time-series data. Linguistic validation was carried out by native-speaking language experts in Bengali, Hindi, Marathi, and Tamil. Each expert has professional experience in medical or technical translation and academic language review. This diverse involvement of experts ensures that MMTClinic is both clinically sound and linguistically accurate.\par
For MCQ type questions:
\begin{itemize}
    \item \textbf{Check Clinical Relevance:} Confirm the question is clinically meaningful and reflects realistic ICU scenarios.
    \item \textbf{Validate Ground Truth Answer:} Ensure the correct option matches the clinical data in the .csv or multivariate plot.
    \item \textbf{Temporal Consistency:} Verify the time-frame referred to in the question aligns with data (e.g., "first 48 hours", "at hour 03:45").
    \item \textbf{No Ambiguity:} Flag any question with ambiguous language or multiple interpretations.
\end{itemize}

For reasoning type questions:
\begin{itemize}
    \item \textbf{Clinical Soundness:} Validate whether the reasoning steps reflect correct pathophysiological logic.
    \item \textbf{Data Justification:} Check if the conclusions are logically derived from provided time-series values and trends.
    \item \textbf{Completeness:} Ensure the reasoning includes all key variables that would influence a real clinical decision.
    \item \textbf{Avoid Overinterpretation:} Ensure no conclusion is drawn beyond what the data supports (e.g., avoid diagnosis without enough evidence).
 \item \textbf{Consistency with SOFA / Mortality Logic:} For subtasks involving SOFA scoring or survival prediction, verify alignment with ICU assessment guidelines.

\end{itemize}

\vspace{-0.3cm}

\subsection{Results}
\label{rs}

In this section, empirical results have been presented in Table \ref{tab1}, \ref{tab2}, \ref{tab3}, \ref{tab4}, \ref{tab5}, \ref{tab6}, \ref{tab10}, \ref{tab11}, \ref{tab111}, \ref{tab112} and \ref{tab113}.

\begin{figure*}[!htbp]
\centering
  \includegraphics[scale=0.6]{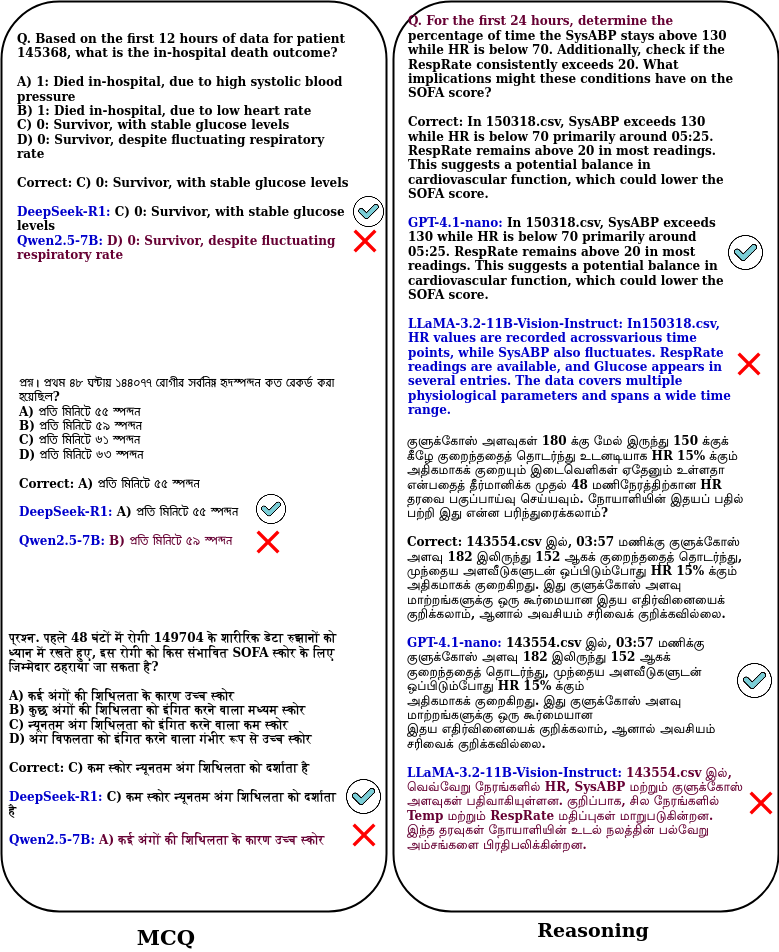}
  \caption{ Example illustration of both MCQ and Reasoning predictions by LLM  }
  \label{fig:example}
\end{figure*}

In Figure \ref{fig:example}, some examples illustration of both MCQ and reasoning question predictions by LLMs have been presented.
\begin{table*}
  \centering
\begin{tabular}{c|c|c|c|c|c}
    \hline
    \textbf{Modality} & \textbf{Type} & \textbf{Task} & \textbf{Language} & \textbf{Model} & \textbf{Accuracy} \\
  \hline
\multicolumn{6}{c}{\textbf{Small LLMs }} \\
\hline
Text + Timeseries & MCQ & Task3 & Bengali & \cellcolor{cyan!25}DeepSeek-R1-Llama-8B & 53.3 \\
Text + Timeseries & MCQ & Task1 & Bengali & \cellcolor{cyan!25}DeepSeek-R1-Llama-8B & 51.5 \\
Text + Timeseries & MCQ & Task2 & Bengali & \cellcolor{cyan!25}DeepSeek-R1-Llama-8B & 52.21 \\

Text + Timeseries & MCQ & Task1 & Bengali & \cellcolor{cyan!25}Mistral-7B-Instruct & 35.05 \\
Text + Timeseries & MCQ & Task2 & Bengali & \cellcolor{cyan!25}Mistral-7B-Instruct & 34.38 \\
Text + Timeseries & MCQ & Task3 & Bengali & \cellcolor{cyan!25}Mistral-7B-Instruct & 23 \\
\hline
  \end{tabular}
  \caption{\label{tab1}
    Model performance across tasks using different Bengali multimodal inputs}
\end{table*}

\begin{table*}
  \centering
  \begin{tabular}{c|c|c|c|c|c}
    \hline
    \textbf{Modality} & \textbf{Type} & \textbf{Task} & \textbf{Language} & \textbf{Model} & \textbf{Accuracy} \\
   \hline
\multicolumn{6}{c}{\textbf{Small LLMs}} \\
\hline
Text + Timeseries & MCQ & Task3 & English & \cellcolor{cyan!25}DeepSeek-R1-Llama-8B & 59.6 \\
Text + Timeseries & MCQ & Task1 & English & \cellcolor{cyan!25}DeepSeek-R1-Llama-8B & 61.24 \\
Text + Timeseries & MCQ & Task2 & English & \cellcolor{cyan!25}DeepSeek-R1-Llama-8B & 58.6 \\

Text + Timeseries & MCQ & Task1 & English & \cellcolor{cyan!25}Mistral-7B-Instruct & 45.99 \\
Text + Timeseries & MCQ & Task2 & English & \cellcolor{cyan!25}Mistral-7B-Instruct & 44.55 \\
Text + Timeseries & MCQ & Task3 & English & \cellcolor{cyan!25}Mistral-7B-Instruct & 32.02 \\
\hline
  \end{tabular}
  \caption{\label{tab2}
    Model performance across tasks using different English multimodal inputs}
\end{table*}

\begin{table*}
  \centering
  \begin{tabular}{c|c|c|c|c|c}
    \hline
    \textbf{Modality} & \textbf{Type} & \textbf{Task} & \textbf{Language} & \textbf{Model} & \textbf{Accuracy} \\
\hline
\multicolumn{6}{c}{\textbf{Small LLMs (≤7B)}} \\
\hline
Text + Timeseries & MCQ & Task3 & Hindi & \cellcolor{cyan!25}DeepSeek-R1-Llama-8B & 55.2 \\
Text + Timeseries & MCQ & Task1 & Hindi & \cellcolor{cyan!25}DeepSeek-R1-Llama-8B & 58.6 \\
Text + Timeseries & MCQ & Task2 & Hindi & \cellcolor{cyan!25}DeepSeek-R1-Llama-8B & 54.1 \\

Text + Timeseries & MCQ & Task1 & Hindi & \cellcolor{cyan!25}Mistral-7B-Instruct & 35.63 \\
Text + Timeseries & MCQ & Task2 & Hindi & \cellcolor{cyan!25}Mistral-7B-Instruct & 40.27 \\
Text + Timeseries & MCQ & Task3 & Hindi & \cellcolor{cyan!25}Mistral-7B-Instruct & 22.86 \\
\hline
  \end{tabular}
  \caption{\label{tab3}
    Model performance across tasks using different Hindi multimodal inputs}
\end{table*}

\begin{table*}
  \centering
  \begin{tabular}{c|c|c|c|c|c}
    \hline
    \textbf{Modality} & \textbf{Type} & \textbf{Task} & \textbf{Language} & \textbf{Model} & \textbf{Accuracy} \\
\hline
\multicolumn{6}{c}{\textbf{Small LLMs (≤7B)}} \\
\hline
Text + Timeseries & MCQ & Task3 & Marathi & \cellcolor{cyan!25}DeepSeek-R1-Llama-8B & 58.7 \\
Text + Timeseries & MCQ & Task1 & Marathi & \cellcolor{cyan!25}DeepSeek-R1-Llama-8B & 59.2 \\
Text + Timeseries & MCQ & Task2 & Marathi & \cellcolor{cyan!25}DeepSeek-R1-Llama-8B & 52.6 \\

Text + Timeseries & MCQ & Task1 & Marathi & \cellcolor{cyan!25}Mistral-7B-Instruct & 33.94 \\
Text + Timeseries & MCQ & Task2 & Marathi & \cellcolor{cyan!25}Mistral-7B-Instruct & 32.86 \\
Text + Timeseries & MCQ & Task3 & Marathi & \cellcolor{cyan!25}Mistral-7B-Instruct & 23.01 \\
\hline
  \end{tabular}
  \caption{\label{tab4}
    Model performance across tasks using different Marathi multimodal inputs}
\end{table*}

\begin{table*}
  \centering
  \begin{tabular}{c|c|c|c|c|c}
\hline
\multicolumn{6}{c}{\textbf{Small LLMs (≤7B)}} \\
\hline
Text + Timeseries & MCQ & Task3 & Tamil & \cellcolor{cyan!25}DeepSeek-R1-Llama-8B & 61.2 \\
Text + Timeseries & MCQ & Task1 & Tamil & \cellcolor{cyan!25}DeepSeek-R1-Llama-8B & 60.2 \\
Text + Timeseries & MCQ & Task2 & Tamil & \cellcolor{cyan!25}DeepSeek-R1-Llama-8B & 56.6 \\

Text + Timeseries & MCQ & Task1 & Tamil & \cellcolor{cyan!25}Mistral-7B-Instruct & 21.48 \\
Text + Timeseries & MCQ & Task2 & Tamil & \cellcolor{cyan!25}Mistral-7B-Instruct & 31.56 \\
Text + Timeseries & MCQ & Task3 & Tamil & \cellcolor{cyan!25}Mistral-7B-Instruct & 25.09 \\
Text + Timeseries & MCQ & Task1 & Tamil & \cellcolor{cyan!25}Qwen2.5-7B & 9.73 \\
\hline
  \end{tabular}
  \caption{\label{tab5}
    Model performance across tasks using different Tamil multimodal inputs}
\end{table*}

\begin{table*}
  \centering
  \begin{tabular}{c|c|c|c|c|c}
    \hline
    \textbf{Modality} & \textbf{Type} & \textbf{Task} & \textbf{Language} & \textbf{Model} & \textbf{Accuracy} \\
\hline
\multicolumn{6}{c}{\textbf{Small to Mid LLMs (≤32B)}} \\
\hline
Text + Timeseries & Reasoning & Task3 & English & \cellcolor{cyan!25}DeepSeek-R1-Llama-8B & 60 \\
Text + Timeseries & Reasoning & Task1 & English & \cellcolor{cyan!25}DeepSeek-R1-Llama-8B & 64.8 \\
Text + Timeseries & Reasoning & Task2 & English & \cellcolor{cyan!25}DeepSeek-R1-Llama-8B & 53.2 \\

Text + Timeseries & Reasoning & Task1 & English & \cellcolor{cyan!25}Mistral-7B-Instruct & 32.05 \\
Text + Timeseries & Reasoning & Task2 & English & \cellcolor{cyan!25}Mistral-7B-Instruct & 31.23 \\
Text + Timeseries & Reasoning & Task3 & English & \cellcolor{cyan!25}Mistral-7B-Instruct & 20 \\
\hline
  \end{tabular}
  \caption{\label{tab6}
    Performance of models on Reasoning tasks using English multimodal inputs}
\end{table*}

\begin{table*}
  \centering
  \begin{tabular}{c|c|c|c|c|c}
    \hline
    \textbf{Modality} & \textbf{Type} & \textbf{Task} & \textbf{Language} & \textbf{Model} & \textbf{Accuracy} \\
\hline
\multicolumn{6}{c}{\textbf{Small to Mid LLMs (≤32B)}} \\
\hline
Text + Timeseries & Reasoning & Task1 & Bengali & \cellcolor{cyan!25}DeepSeek-R1-Llama-8B & 76.6 \\
Text + Timeseries & Reasoning & Task2 & Bengali & \cellcolor{cyan!25}DeepSeek-R1-Llama-8B & 54.8 \\
Text + Timeseries & Reasoning & Task3 & Bengali & \cellcolor{cyan!25}DeepSeek-R1-Llama-8B & 64.6 \\

Text + Timeseries & Reasoning & Task1 & Bengali & \cellcolor{cyan!25}Mistral-7B-Instruct & 30.0 \\
Text + Timeseries & Reasoning & Task2 & Bengali & \cellcolor{cyan!25}Mistral-7B-Instruct & 25.3 \\
Text + Timeseries & Reasoning & Task3 & Bengali & \cellcolor{cyan!25}Mistral-7B-Instruct & 18.92 \\
\hline
  \end{tabular}
  \caption{\label{tab7}
    Performance of models on Reasoning tasks using Bengali multimodal inputs}
\end{table*}

\begin{table*}
  \centering
  \begin{tabular}{c|c|c|c|c|c}
    \hline
    \textbf{Modality} & \textbf{Type} & \textbf{Task} & \textbf{Language} & \textbf{Model} & \textbf{Accuracy} \\
 \hline
\multicolumn{6}{c}{\textbf{Small to Mid LLMs (≤32B)}} \\
\hline
Text + Timeseries & Reasoning & Task2 & Hindi & \cellcolor{cyan!25}DeepSeek-R1-Llama-8B & 57.6 \\
Text + Timeseries & Reasoning & Task3 & Hindi & \cellcolor{cyan!25}DeepSeek-R1-Llama-8B & 66.6 \\
Text + Timeseries & Reasoning & Task1 & Hindi & \cellcolor{cyan!25}DeepSeek-R1-Llama-8B & 75.6 \\

Text + Timeseries & Reasoning & Task1 & Hindi & \cellcolor{cyan!25}Mistral-7B-Instruct & 28.6 \\
Text + Timeseries & Reasoning & Task2 & Hindi & \cellcolor{cyan!25}Mistral-7B-Instruct & 24.7 \\
Text + Timeseries & Reasoning & Task3 & Hindi & \cellcolor{cyan!25}Mistral-7B-Instruct & 17.43 \\
\hline
  \end{tabular}
  \caption{\label{tab8}
    Performance of models on Reasoning tasks using Hindi multimodal inputs}
\end{table*}
\begin{table*}
  \centering
  \begin{tabular}{c|c|c|c|c|c}
    \hline
    \textbf{Modality} & \textbf{Type} & \textbf{Task} & \textbf{Language} & \textbf{Model} & \textbf{Accuracy} \\
\hline
\multicolumn{6}{c}{\textbf{Small to Mid LLMs (≤32B)}} \\
\hline
Text + Timeseries & Reasoning & Task1 & Marathi & \cellcolor{cyan!25}DeepSeek-R1-Llama-8B & 70.6 \\
Text + Timeseries & Reasoning & Task2 & Marathi & \cellcolor{cyan!25}DeepSeek-R1-Llama-8B & 51.4 \\
Text + Timeseries & Reasoning & Task3 & Marathi & \cellcolor{cyan!25}DeepSeek-R1-Llama-8B & 60 \\

Text + Timeseries & Reasoning & Task1 & Marathi & \cellcolor{cyan!25}Mistral-7B-Instruct & 26.5 \\
Text + Timeseries & Reasoning & Task2 & Marathi & \cellcolor{cyan!25}Mistral-7B-Instruct & 21.3 \\
Text + Timeseries & Reasoning & Task3 & Marathi & \cellcolor{cyan!25}Mistral-7B-Instruct & 17.45 \\
\hline
  \end{tabular}
  \caption{\label{tab9}
    Performance of models on Reasoning tasks using Marathi multimodal inputs}
\end{table*}
\begin{table*}
  \centering
  \begin{tabular}{c|c|c|c|c|c}
    \hline
    \textbf{Modality} & \textbf{Type} & \textbf{Task} & \textbf{Language} & \textbf{Model} & \textbf{Accuracy} \\
\hline
\multicolumn{6}{c}{\textbf{Small to Mid LLMs (≤32B)}} \\
\hline
Text + Timeseries & Reasoning & Task1 & Tamil & \cellcolor{cyan!25}DeepSeek-R1-Llama-8B & 80.8 \\
Text + Timeseries & Reasoning & Task3 & Tamil & \cellcolor{cyan!25}DeepSeek-R1-Llama-8B & 65.2 \\
Text + Timeseries & Reasoning & Task2 & Tamil & \cellcolor{cyan!25}DeepSeek-R1-Llama-8B & 56.2 \\

Text + Timeseries & Reasoning & Task1 & Tamil & \cellcolor{cyan!25}Mistral-7B-Instruct & 23.87 \\
Text + Timeseries & Reasoning & Task2 & Tamil & \cellcolor{cyan!25}Mistral-7B-Instruct & 23.43 \\
Text + Timeseries & Reasoning & Task3 & Tamil & \cellcolor{cyan!25}Mistral-7B-Instruct & 18.2 \\
\hline
  \end{tabular}
  \caption{\label{tab10}
    Performance of models on Reasoning tasks using Tamil multimodal inputs}
\end{table*}
\begin{table*}[htbp]
\centering
\scriptsize  
\setlength{\tabcolsep}{15.5pt}  
\renewcommand{\arraystretch}{0.9}  
  \begin{tabular}{c|c|c|c|c|c}
    \hline
    \textbf{Modality} & \textbf{Type} & \textbf{Task} & \textbf{Language} & \textbf{Model} & \textbf{Accuracy} \\
    \hline
\rowcolor{cyan!15}Text + Timeseries + Image & MCQ & Task1 & Bengali & Llama-3.2-11B-Vision-Instruct & 30.05 \\
\rowcolor{cyan!15}Text + Timeseries + Image & MCQ & Task2 & Bengali & Llama-3.2-11B-Vision-Instruct & 39.77 \\
\rowcolor{cyan!15}Text + Timeseries + Image & MCQ & Task3 & Bengali & Llama-3.2-11B-Vision-Instruct & 27.81 \\
\rowcolor{cyan!15}Text + Timeseries + Image & MCQ & Task1 & English & Llama-3.2-11B-Vision-Instruct & 34.48 \\
\rowcolor{cyan!15}Text + Timeseries + Image & MCQ & Task2 & English & Llama-3.2-11B-Vision-Instruct & 39.26 \\
\rowcolor{cyan!15}Text + Timeseries + Image & MCQ & Task3 & English & Llama-3.2-11B-Vision-Instruct & 30.96 \\
\rowcolor{cyan!15}Text + Timeseries + Image & MCQ & Task1 & Hindi & Llama-3.2-11B-Vision-Instruct & 25.53 \\
\hline
  \end{tabular}
  \caption{\label{tab11}
    Performance of models on MCQ tasks with Text + Timeseries + Image inputs across multiple languages}
\end{table*}

\begin{table*}[htbp]
\centering
\scriptsize  
\setlength{\tabcolsep}{15.5pt}  
\renewcommand{\arraystretch}{0.9}  
\begin{tabular}{c|c|c|c|c|c}
\hline
\textbf{Modality} & \textbf{Type} & \textbf{Task} & \textbf{Language} & \textbf{Model} & \textbf{Accuracy} \\
\hline
\rowcolor{cyan!15}Text + Timeseries  & MCQ & Task1 & Bengali & Llama-3.2-11B-Vision-Instruct & 33.07 \\
\rowcolor{cyan!15}Text + Timeseries  & MCQ & Task2 & Bengali & Llama-3.2-11B-Vision-Instruct & 44.61 \\
\rowcolor{cyan!15}Text + Timeseries  & MCQ & Task3 & Bengali & Llama-3.2-11B-Vision-Instruct & 31.74 \\
\rowcolor{cyan!15}Text + Timeseries  & MCQ & Task1 & English & Llama-3.2-11B-Vision-Instruct & 39.07 \\
\rowcolor{cyan!15}Text + Timeseries  & MCQ & Task2 & English & Llama-3.2-11B-Vision-Instruct & 43.98 \\
\rowcolor{cyan!15}Text + Timeseries  & MCQ & Task3 & English & Llama-3.2-11B-Vision-Instruct & 34.92 \\
\rowcolor{cyan!15}Text + Timeseries  & MCQ & Task1 & Hindi & Llama-3.2-11B-Vision-Instruct & 29.23 \\
\hline
\end{tabular}

\caption{\label{tab111}Performance of Few-Shot prompting for MCQ tasks across different languages and models}
\end{table*}

\begin{table*}[htbp]
\centering
\scriptsize
\setlength{\tabcolsep}{15.5pt}
\renewcommand{\arraystretch}{0.9}
\begin{tabular}{c|c|c|c|c|c}
\hline
\textbf{Modality} & \textbf{Type} & \textbf{Task} & \textbf{Language} & \textbf{Model} & \textbf{Accuracy} \\
\hline
\rowcolor{cyan!15}Text + Timeseries & MCQ & Task1 & Bengali & Llama-3.2-11B-Vision-Instruct & 38.63 \\
\rowcolor{cyan!15}Text + Timeseries & MCQ & Task2 & Bengali & Llama-3.2-11B-Vision-Instruct & 50.72 \\
\rowcolor{cyan!15}Text + Timeseries & MCQ & Task3 & Bengali & Llama-3.2-11B-Vision-Instruct & 38.39 \\
\rowcolor{cyan!15}Text + Timeseries & MCQ & Task1 & English & Llama-3.2-11B-Vision-Instruct & 45.51 \\
\rowcolor{cyan!15}Text + Timeseries & MCQ & Task2 & English & Llama-3.2-11B-Vision-Instruct & 50.28 \\
\rowcolor{cyan!15}Text + Timeseries & MCQ & Task3 & English & Llama-3.2-11B-Vision-Instruct & 41.01 \\
\rowcolor{cyan!15}Text + Timeseries & MCQ & Task1 & Hindi & Llama-3.2-11B-Vision-Instruct & 35.30 \\
\hline
\end{tabular}

\caption{\label{tab112}Performance of CoT prompting for MCQ tasks across different languages and models}
\end{table*}
\begin{table*}[htbp]
\centering
\scriptsize
\setlength{\tabcolsep}{15.5pt}
\renewcommand{\arraystretch}{0.8}
\begin{tabular}{c|c|c|c|c|c}
\hline
\textbf{Modality} & \textbf{Type} & \textbf{Task} & \textbf{Language} & \textbf{Model} & \textbf{Accuracy} \\
\hline
\rowcolor{cyan!15}Text + Timeseries + Image & MCQ & Task1 & Bengali & Llama-3.2-11B-Vision-Instruct & 38.26 \\
\rowcolor{cyan!15}Text + Timeseries + Image & MCQ & Task2 & Bengali & Llama-3.2-11B-Vision-Instruct & 52.51 \\
\rowcolor{cyan!15}Text + Timeseries + Image & MCQ & Task3 & Bengali & Llama-3.2-11B-Vision-Instruct & 36.60 \\
\rowcolor{cyan!15}Text + Timeseries + Image & MCQ & Task1 & English & Llama-3.2-11B-Vision-Instruct & 47.00 \\
\rowcolor{cyan!15}Text + Timeseries + Image & MCQ & Task2 & English & Llama-3.2-11B-Vision-Instruct & 50.72 \\
\rowcolor{cyan!15}Text + Timeseries + Image & MCQ & Task3 & English & Llama-3.2-11B-Vision-Instruct & 41.71 \\
\rowcolor{cyan!15}Text + Timeseries + Image & MCQ & Task1 & Hindi & Llama-3.2-11B-Vision-Instruct & 33.38 \\
\hline
\end{tabular}

\caption{\label{tab113}Few-Shot Accuracy for reasoning tasks across different languages and models using Text + Timeseries + Image modality}
\end{table*}

\begin{table*}[htbp]
\centering
\scriptsize
\setlength{\tabcolsep}{15.5pt}
\renewcommand{\arraystretch}{0.9}
\begin{tabular}{c|c|c|c|c|c}
\hline
\textbf{Modality} & \textbf{Type} & \textbf{Task} & \textbf{Language} & \textbf{Model} & \textbf{Accuracy} \\
\hline
\rowcolor{cyan!15}Text + Timeseries + Image & MCQ & Task1 & Bengali & Llama-3.2-11B-Vision-Instruct & 45.63\\
\rowcolor{cyan!15}Text + Timeseries + Image & MCQ & Task2 & Bengali & Llama-3.2-11B-Vision-Instruct & 58.76 \\
\rowcolor{cyan!15}Text + Timeseries + Image & MCQ & Task3 & Bengali & Llama-3.2-11B-Vision-Instruct & 39.82 \\
\rowcolor{cyan!15}Text + Timeseries + Image & MCQ & Task1 & English & Llama-3.2-11B-Vision-Instruct & 48.82 \\
\rowcolor{cyan!15}Text + Timeseries + Image & MCQ & Task2 & English & Llama-3.2-11B-Vision-Instruct & 55.39 \\
\rowcolor{cyan!15}Text + Timeseries + Image & MCQ & Task3 & English & Llama-3.2-11B-Vision-Instruct & 47.87 \\
\rowcolor{cyan!15}Text + Timeseries + Image & MCQ & Task1 & Hindi & Llama-3.2-11B-Vision-Instruct & 37.72 \\
\hline
\end{tabular}

\caption{\label{tab114} CoT Accuracy for reasoning tasks across different languages and models using Text + Timeseries + Image modality}
\end{table*}

\begin{table*}[htbp]
\centering
\scriptsize  
\setlength{\tabcolsep}{15.5pt}  
\renewcommand{\arraystretch}{0.9}  
\begin{tabular}{c|c|c|c|c|c}
\hline
\textbf{Modality} & \textbf{Type} & \textbf{Task} & \textbf{Language} & \textbf{Model} & \textbf{Accuracy} \\
\hline
\rowcolor{cyan!15}Text + Timeseries + Image & Reasoning & Task1 & Bengali & Llama-3.2-11B-Vision-Instruct & 29.25 \\
\rowcolor{cyan!15}Text + Timeseries + Image & Reasoning & Task2 & Bengali & Llama-3.2-11B-Vision-Instruct & 33.34 \\
\rowcolor{cyan!15}Text + Timeseries + Image & Reasoning & Task3 & Bengali & Llama-3.2-11B-Vision-Instruct & 21.31 \\
\rowcolor{cyan!15}Text + Timeseries + Image & Reasoning & Task1 & English & Llama-3.2-11B-Vision-Instruct & 30.12 \\
\rowcolor{cyan!15}Text + Timeseries + Image & Reasoning & Task2 & English & Llama-3.2-11B-Vision-Instruct & 37.19 \\
\rowcolor{cyan!15}Text + Timeseries + Image & Reasoning & Task3 & English & Llama-3.2-11B-Vision-Instruct & 23.8 \\
\rowcolor{cyan!15}Text + Timeseries + Image & Reasoning & Task1 & Hindi & Llama-3.2-11B-Vision-Instruct & 23.12 \\
\hline
\end{tabular}
\caption{\label{tab:reasoning}
Performance of models on *Reasoning* tasks with Text + Timeseries + Image inputs across multiple languages. }
\end{table*}
\onecolumn

\subsection{Prompts}
\label{pmt}
\noindent\textbf{A. Prompt for MCQ-QAs based on Text and Time Series:}
\begin{lstlisting}[caption={ICU Task-based MCQ Generation Prompt}]
You are a data assistant for ICU time-series analysis.

You have access to:
- (*@\textbf{\{patient\_id\}.csv}@*): (Time, Parameter, Value)
- (*@\textbf{Outcomes-a.csv}@*): Patient-level in-hospital outcomes

Focus only on: HR, DiasABP, SysABP, RespRate, Temp, Glucose

Generate *2 distinct MCQs* for each of the following tasks:

---

(*@\textbf{Subtask 1: Predicting In-hospital Death}@*)  
Answer is directly available from Outcomes-a.csv.  
Base questions on early (6–12h) stats or threshold values.

\textit{Example format:}  
Q. Based on vitals from the first 12 hours, what was the patient's in-hospital outcome?  
A) Discharged to home  
B) Stable with ICU stay  
C) (*@\textbf{Died in-hospital}@*)  
D) Recovered without complications  
Correct Answer: C — Died in-hospital (Outcomes-a.csv = 1)

---

(*@\textbf{Subtask 2: Forecasting Heart Rate (HR)}@*)  
Use 48-hour HR trends from {patient_id}.csv to compute mean/max/min.

\textit{Example format:}  
Q. What was the minimum HR observed in the first 48 hours?  
A) 85 bpm  
B) 98 bpm  
C) (*@\textbf{76 bpm}@*)  
D) 101 bpm  
Correct Answer: C — 76 bpm (observed in file)

---

(*@\textbf{Subtask 3: Estimating SOFA Score (Hypothetical)}@*)  
Use parameter trends to infer plausible SOFA severity level.  
Base answers on clinical patterns from first 48h.

\textit{Example format:}  
Q. Based on sustained hypotension and respiratory distress, what is the likely SOFA status?  
A) Normal physiology  
B) Mild dysfunction (SOFA 1–3)  
C) (*@\textbf{Moderate dysfunction (SOFA 4–6)}@*)  
D) Multi-organ failure (SOFA >10)  
Correct Answer: C — Consistent with observed multi-parameter abnormality

---

(*@\textbf{Sample Data}@*): First 20 rows from \{patient\_id\}.csv  
(*@\texttt{{patient\_df.head(20).to\_string(index=False)}}@*)

Outcome from Outcomes-a.csv: **{in_hospital_death}**

Now generate **6 MCQs (2 per subtask)** with:
- 4 options (A–D)
- 1 clearly correct answer (justified by data or outcome)
- 3 plausible distractors
- No reuse of prior templates or examples
\end{lstlisting}

\noindent\textbf{B.Prompt for MCQ-QAs based on Text, Time Series, and Image:}
\begin{lstlisting}[ caption={Multitask ICU MCQ Generation Prompt}]
You are a data assistant for ICU time-series analysis.

You have access to:
- <(*@\textbf{\{patient\_id\}.csv}@*)>: (Time, Parameter, Value)
- <(*@\textbf{\{patient\_id\}_main.png}@*)>: Line plots of vitals
- <(*@\textbf{Outcomes-a.csv}@*)>: In-hospital outcomes

Focus only on these parameters: HR, DiasABP, SysABP, RespRate, Temp, Glucose

Generate exactly **2 multiple-choice questions (MCQs)** for each subtask:

---

(*@\textbf{Subtask 1: Predicting In-hospital death}@*)  
Use the first 6–12 hours of time-series data and compare with outcome in Outcomes-a.csv.  

\textbf{Example format:}  
Q. Based on the patient's early vitals, what is the likely in-hospital outcome?  
A) Survived with stable vitals  
B) Unknown  
C) (*@\textbf{Died in-hospital}@*)  
D) Transferred to step-down  
Correct Answer: C — Died in-hospital (Outcomes-a.csv = 1)

---

(*@\textbf{Subtask 2: Forecasting HR during first 48 hours}@*)  
Use HR trends from the CSV to calculate mean, max, min, or specific time-window comparisons.

\textbf{Example format:}  
Q. What was the maximum HR recorded during the first 48 hours?  
A) 102 bpm  
B) (*@\textbf{110 bpm}@*)  
C) 95 bpm  
D) 88 bpm  
Correct Answer: B — 110 bpm (observed from {patient_id}.csv)

---

(*@\textbf{Subtask 3: Estimating SOFA score severity}@*)  
Use physiological deviations to infer plausible SOFA-related conditions. Ground answer in trends from 48h data.

\textbf{Example format:}  
Q. Based on sustained low blood pressure and elevated respiration, what is the plausible SOFA status?  
A) No organ dysfunction  
B) Minor dysfunction (SOFA < 2)  
C) (*@\textbf{Moderate dysfunction (SOFA 4–6)}@*)  
D) Severe failure (SOFA > 10)  
Correct Answer: C — Moderate dysfunction inferred from BP and RespRate patterns

---

Below is a sample from {patient_id}'s ICU time-series data:  
(*@\texttt{{patient\_df.head(20).to\_string(index=False)}}@*)

Outcome: In-hospital death = (*@\textbf{\{in\_hospital\_death\}}@*)  
(*@\textbf{Image: \{patient\_id\}_main.png}@*)

---

Now generate exactly **6 MCQs** (2 per subtask), with:
- 4 answer options per question (A–D)
- One correct and clearly supported/inferred answer
- Three plausible but incorrect distractors
- No reuse of examples or prior questions
\end{lstlisting}

\noindent\textbf{C. Prompt for Reasoning-QAs based on Text and Time Series:}
\begin{lstlisting}[ caption={ICU Reasoning QA Generation Prompt}]
You are a specialized data analyst for ICU time-series.

You have access to:
- (*@\textbf{\{patient\_id\}.csv}@*): Contains (Time, Parameter, Value)
- (*@\textbf{Outcomes-a.csv}@*): Contains patient outcomes

Use only these parameters: HR, DiasABP, SysABP, RespRate, Temp, Glucose

Generate **one reasoning-style open-ended question per task**:

- Subtask 1: Predicting in-hospital death (0: survived, 1: died)
- Subtask 2: Forecasting HR using first 48h of HR time-series
- Subtask 3: Predicting SOFA score using multivariate ICU data

Each question must:
1. Use at least two parameters or timestamps
2. Require multi-hop numeric logic (trends, comparisons, averages, thresholds)
3. Include a brief answer using only (*@\textbf{\{patient\_id\}.csv}@*) and (*@\textbf{Outcomes-a.csv}@*)

---

(*@\textbf{Example Format}@*)

\textbf{Subtask 1}  
Question:  
Check if the patient has ≥3 episodes (4h each) of hypotension (SysABP < 90, DiasABP < 60) and tachycardia (HR > 100) in the first 48h. Based on these, did the patient survive?

Answer:  
Data shows ≥3 qualifying episodes. According to Outcomes-a.csv, in-hospital death = {in_hospital_death}.

---

\textbf{Subtask 2}  
Question:  
Does HR increase by >20% within 8h of Glucose >180 (compared to the 4h average before)? If yes, is this an early warning sign?

Answer:  
Yes, at least two instances show this pattern in the 48h window.

---

\textbf{Subtask 3}  
Question:  
Average DiasABP and SysABP, RespRate >25 (hourly count), and Temp <36°C occurrences. Do these patterns indicate elevated SOFA risk?

Answer:  
DiasABP is low, RespRate >25 in 15+ hours, Temp <36°C observed. SOFA score in Outcomes-a.csv = {sofa_score}.

---

Patient snapshot (top 20 rows of \{patient\_id\}.csv):  
(*@\texttt{{patient\_df.head(20).to\_string(index=False)}}@*)

Outcome: In-hospital death = (*@\textbf{\{in\_hospital\_death\}}@*)  
(*@\textbf{SOFA score = \{sofa\_score\}}@*) if available

Now generate one new question and answer per subtask (not reused from above).
\end{lstlisting}

\noindent\textbf{D. Prompt for Reasoning-QAs based on Text, Time Series, and Image:}
\begin{lstlisting}[ caption={Reasoning QA Generation Prompt}]
You are a data assistant for ICU time-series reasoning.

You have access to:
- <"{patient_id}.csv">: (Time, Parameter, Value) — ICU vitals
- <"{patient_id}_main.png">: line plots of HR, DiasABP, SysABP, RespRate, Temp, Glucose
- <"Outcomes-a.csv">: patient outcome data

Only consider these parameters from {patient_id}.csv:
HR, DiasABP, SysABP, RespRate, Temp, Glucose

Generate **two reasoning questions per subtask** for:

- Subtask 1: Predicting In-hospital death (0 = survived, 1 = died)
- Subtask 2: Forecasting HR based on first 48h HR data
- Subtask 3: Predicting SOFA score from multivariate time-series

Each question must:
1. Involve at least two parameters or timestamps
2. Require multi-hop numeric logic (e.g., trends, comparisons, averages)
3. Include a brief **Answer** using only data from the three files

---

(*@\textbf{Example Format}@*)

\textbf{Subtask 1}  
Question:  
Check for hypotension (SysABP < 90, DiasABP < 60) with HR > 100 for 3+ distinct 4h periods. Based on this, what does <"Outcomes-a.csv"> report for in-hospital death?

Answer:  
{patient_id}.csv shows these conditions met 3+ times. In-hospital_death from Outcomes-a.csv is {in_hospital_death}.

---

\textbf{Subtask 2}  
Question:  
Does HR rise >20% within 8h after Glucose >180 compared to the 4h average before? If so, does this suggest early deterioration?

Answer:  
Yes, in 2+ instances, HR spikes >20% post-glucose. Indicates deterioration.

---

\textbf{Subtask 3}  
Question:  
Avg DiasABP and SysABP in first 48h + count RespRate >25 + any Temp <36°C. Do these support high SOFA score?

Answer:  
DiasABP low, RespRate >25 in 15+ hours, Temp <36°C twice. SOFA in Outcomes-a.csv is {sofa_score}.

---

Sample from {patient_id}.csv (top 20 rows):  
{patient_df.head(20).to_string(index=False)}

In-hospital death: **{in_hospital_death}**  
{f"SOFA score: **{sofa_score}**" if sofa_score is not None else ""}

---

Now generate two novel reasoning-style Q&A pairs per subtask using the data. Do not reuse previous questions.
\end{lstlisting}

\noindent\textbf{E. Prompt for Reasoning-QA evaluation based on Text and Time Series:}
\begin{lstlisting}[ caption={ICU Time-Series Reasoning Prompt}]
You are a medical expert in ICU time-series interpretation.

Input:
- Patient CSV: (*@\textbf{\{PATIENT\_DF\}}@*)
Vitals: HR, DiasABP, SysABP, RespRate, Temp, Glucose

Your task:
1. Analyze the data
2. Answer the reasoning question
3. Return `1` if data clearly supports a conclusion
4. Return `0` if data is inconclusive

Task: (*@\textbf{\{TASK\}}@*)

---

Patient Data:
(*@\textbf{\{PATIENT\_DF\}}@*)

Question:
(*@\textbf{\{QUESTION\}}@*)

---

Instructions:
- Use only the given data
- No external medical knowledge
- Return only: `1` or `0`
- (*@\textbf{No explanations or extra text}@*)
\end{lstlisting}

\noindent\textbf{F. Prompt for Reasoning-QA evaluation based on Text, Time Series, and Image:}
\begin{lstlisting}[ caption={Multimodal Reasoning Prompt}]
You are a medical expert in ICU time-series and signal plot interpretation.

Inputs:
- CSV file: (*@\textbf{\{PATIENT\_DF\}}@*)
- Plot image: (*@\textbf{\{IMAGE\}}@*)
Parameters: HR, DiasABP, SysABP, RespRate, Temp, Glucose

Your task:
1. Review the data and image
2. Assess the reasoning question
3. Return `1` if evidence supports a clear conclusion
4. Return `0` if evidence is insufficient

Task: (*@\textbf{\{TASK\}}@*)

---

Patient Data:
(*@\textbf{\{PATIENT\_DF\}}@*)

Image:
(*@\textbf{\{IMAGE\}}@*)

Question:
(*@\textbf{\{QUESTION\}}@*)

---

Instructions:
- Use only the given data and image
- Do not apply external medical knowledge
- Return only: `1` or `0`
- (*@\textbf{No explanations or additional output}@*)
\end{lstlisting}

\noindent\textbf{G. Prompt for MCQ-QA evaluation based on Text and Time Series:}
\begin{lstlisting}[ caption={ICU Time-Series MCQ Prompt}]
You are an expert in interpreting ICU time-series data.

The patient data is in CSV format with one row per timestamp.
Parameters: HR, DiasABP, SysABP, RespRate, Temp, Glucose

Use only the data below to answer the MCQ.
Return (*@\textbf{only one letter}@*): A, B, C, or D.
Do not explain or output anything else.

Task: (*@\textbf{\{TASK\}}@*)

---

Patient Data:
(*@\textbf{\{PATIENT\_DF\}}@*)

Question:
(*@\textbf{\{QUESTION\}}@*)

Options:
(*@\textbf{\{OPTIONS\}}@*)
\end{lstlisting}

\noindent\textbf{H. Prompt for MCQ-QA evaluation based on Text, Time Series and Image:}
\begin{lstlisting}[caption={Multimodal ICU Prompt}]
You are an expert in interpreting ICU time-series data.

Given patient vitals in CSV format and a corresponding image, 
use the information to answer the following multiple-choice question.

Available parameters:
- HR, DiasABP, SysABP, RespRate, Temp, Glucose

Use only the data: (*@\textbf{\{PATIENT\_DF\}}@*) and image: (*@\textbf{\{IMAGE\}}@*)

The question pertains to: (*@\textbf{Task \{TASK\}}@*)

---

Patient Data:
(*@\textbf{\{PATIENT\_DF\}}@*)

Image:
(*@\textbf{\{IMAGE\}}@*)

Question:
(*@\textbf{\{QUESTION\}}@*)

Options:
(*@\textbf{\{OPTIONS\}}@*)

Return only one letter: A, B, C, or D.
\end{lstlisting}

\noindent\textbf{I. Few-shot prompt for MCQ-QA evaluation based on Text and Time Series:}
\begin{lstlisting}[ caption={ICU Time-Series MCQ Prompt}]
You are an expert in interpreting ICU time-series data.

You are given a CSV file containing vital signs for a patient.  
Each row corresponds to one timestamp, with parameters:
- HR (Heart Rate)
- DiasABP (Diastolic Arterial BP)
- SysABP (Systolic Arterial BP)
- RespRate (Respiration Rate)
- Temp (Body Temperature)
- Glucose (Blood Glucose Level)

---

(*@\textbf{Example 1:}@*)  
Question: (*@\texttt{\{QUESTION1\}}@*)  
Options:  
(*@\texttt{\{OPTION1\}}@*)  
Answer: (*@\texttt{\{ANSWER1\}}@*)

---

(*@\textbf{Example 2:}@*)  
Question: (*@\texttt{\{QUESTION2\}}@*)  
Options:  
(*@\texttt{\{OPTION2\}}@*)  
Answer: (*@\texttt{\{ANSWER2\}}@*)

---

Use only the provided data to answer the following multiple-choice question.  
Strictly return (*@\textbf{only the letter}@*): A, B, C, or D.  
(*@\textbf{No explanation. No extra text.}@*)

---

Task: (*@\textbf{\{TASK\}}@*)

Patient Data:  
(*@\texttt{\{PATIENT\_DF\}}@*)

---

Question:  
(*@\texttt{\{QUESTION\}}@*)

Options:  
(*@\texttt{\{OPTIONS\}}@*)

Answer with only one letter: A, B, C, or D
\end{lstlisting}

\noindent\textbf{J. CoT prompt for MCQ-QA evaluation based on Text and Time Series:}

\begin{lstlisting}[caption={Clinical Reasoning Prompt with Step-by-Step Analysis}]
You are a clinical decision support system specializing in interpreting ICU time-series physiological data to assist with evidence-based reasoning.

Your task is to analyze patient vital trends and apply step-by-step clinical logic to answer a multiple-choice question.

The time-series dataset includes:
- HR: Heart Rate (bpm)
- SysABP: Systolic Arterial BP (mmHg)
- DiasABP: Diastolic Arterial BP (mmHg)
- RespRate: Respiratory Rate (breaths/min)
- Temp: Body Temperature (°C)
- Glucose: Blood Glucose Level (mg/dL)

You will be given a clinical task and patient data. Use the structured reasoning format below.

---

(*@\textbf{Example 1}@*)  
Task: Clinical deterioration prediction  
Question: (*@\texttt{\{QUESTION1\}}@*)  
Options:  
(*@\texttt{\{OPTION1\}}@*)  
Answer Reasoning:  
Abnormal HR and RespRate with falling BP suggest decompensation.  
The likely interpretation is: (*@\textbf{\{ANSWER1\}}@*)

---

(*@\textbf{Example 2}@*)  
Task: Detection of glycemic instability  
Question: (*@\texttt{\{QUESTION2\}}@*)  
Options:  
(*@\texttt{\{OPTION2\}}@*)  
Answer Reasoning:  
Glucose fluctuation with temp/HR change implies poor glycemic control.  
The correct conclusion is: (*@\textbf{\{ANSWER2\}}@*)

---

(*@\textbf{Now analyze the following ICU case using this reasoning template}@*)

Task: (*@\textbf{\{TASK\}}@*)

---

Patient Data:  
(*@\texttt{\{PATIENT\_DF\}}@*)

---

Question:  
(*@\texttt{\{QUESTION\}}@*)

---

Options (choose one):  
(*@\texttt{\{OPTIONS\}}@*)

---

Step-by-Step Reasoning:  
(Interpret trends, flag abnormalities, and link findings to task context.)

Final Answer: (*@\textbf{A, B, C, or D}@*)
\end{lstlisting}
\noindent\textbf{K. Few-shot prompt for MCQ-QA evaluation based on Text, Time Series and Image:}
\begin{lstlisting}[ caption={ICU Multimodal MCQ Prompt (Time-Series + Image)}]
You are an expert in interpreting ICU time-series and image-based physiological data.

You are provided with:
- Vital signs CSV data for a patient (timestamped time-series)
- A corresponding line plot image visualizing parameter trends

Available parameters include:
- HR (Heart Rate)
- DiasABP (Diastolic Arterial BP)
- SysABP (Systolic Arterial BP)
- RespRate (Respiratory Rate)
- Temp (Body Temperature)
- Glucose (Blood Glucose Level)

---

(*@\textbf{Example 1}@*)  
Question: (*@\texttt{\{QUESTION1\}}@*)  
Options:  
(*@\texttt{\{OPTION1\}}@*)  
Answer: (*@\textbf{\{ANSWER1\}}@*)

---

(*@\textbf{Example 2}@*)  
Question: (*@\texttt{\{QUESTION2\}}@*)  
Options:  
(*@\texttt{\{OPTION2\}}@*)  
Answer: (*@\textbf{\{ANSWER2\}}@*)

---

Use only the provided time-series data and image to answer the MCQ below.

Task: (*@\textbf{\{TASK\}}@*)

---

Patient Data:  
(*@\texttt{\{PATIENT\_DF\}}@*)

Image:  
(*@\texttt{\{IMAGE\}}@*)

---

Question:  
(*@\texttt{\{QUESTION\}}@*)

Options:  
(*@\texttt{\{OPTIONS\}}@*)

Answer strictly with one letter only: (*@\textbf{A, B, C, or D}@*)
\end{lstlisting}

\noindent\textbf{L. CoT prompt for MCQ-QA evaluation based on Text, Time Series and Image:}
\begin{lstlisting}[ caption={Multimodal Clinical Reasoning Prompt (Time-Series + Image)}]
You are a clinical decision support system specializing in analyzing ICU time-series data and corresponding line plot images.  
Your task is to evaluate physiological patterns and reason through multiple-choice questions using clinical logic.

Input:
- ICU patient vital signs (CSV time-series)
- Line plot image showing trends in HR, SysABP, DiasABP, RespRate, Temp, Glucose

---

(*@\textbf{Example 1}@*)  
Task: Predicting cardiovascular instability  
Question: (*@\texttt{\{QUESTION1\}}@*)  
Options:  
(*@\texttt{\{OPTION1\}}@*)  
Answer Reasoning:  
The image shows declining BP; CSV confirms tachycardia and rising RespRate. Together, this suggests early shock.  
Correct Answer: (*@\textbf{\{ANSWER1\}}@*)

---

(*@\textbf{Example 2}@*)  
Task: Detecting metabolic abnormality  
Question: (*@\texttt{\{QUESTION2\}}@*)  
Options:  
(*@\texttt{\{OPTION2\}}@*)  
Answer Reasoning:  
Image shows glucose swings and dips; CSV aligns with temp spikes, indicating metabolic stress.  
Correct Answer: (*@\textbf{\{ANSWER2\}}@*)

---

(*@\textbf{Now apply the same clinical reasoning to the case below}@*)

Task: (*@\textbf{\{TASK\}}@*)

---

Patient Data (CSV):  
(*@\texttt{\{PATIENT\_DF\}}@*)

Image (Line Plot):  
(*@\texttt{\{IMAGE\}}@*)

---

Question:  
(*@\texttt{\{QUESTION\}}@*)

Options (choose one):  
(*@\texttt{\{OPTIONS\}}@*)

---

Step-by-Step Reasoning:  
(Examine vitals + image. Identify abnormalities, infer clinical implications, and relate them to the question.)

Final Answer: (*@\textbf{A, B, C, or D}@*)
\end{lstlisting}

\noindent\textbf{M. Few-shot prompt for Reasoning-QA evaluation based on Text, Time Series and Image:}
\begin{lstlisting}[ caption={Binary Classification Prompt (Multimodal Clinical Reasoning)}]
You are a medical expert in interpreting ICU time-series data and physiological signal plots.

You are given:
- ICU patient data in CSV format
- A time-series line plot image of vital parameters

Parameters include:
- HR (Heart Rate)
- DiasABP (Diastolic Arterial BP)
- SysABP (Systolic Arterial BP)
- RespRate (Respiration Rate)
- Temp (Body Temperature)
- Glucose (Blood Glucose Level)

You must:
1. Analyze patient data and line plot
2. Evaluate the reasoning question
3. Return `1` if the data supports the conclusion
4. Return `0` if the data is inconclusive

---

(*@\textbf{Example 1}@*)  
Task: (*@\texttt{\{EX\_TASK\_1\}}@*)  
Patient Data (CSV):  
(*@\texttt{\{EX\_PATIENT\_DF\_1\}}@*)  
Vital Sign Image:  
(*@\texttt{\{EX\_IMAGE\_1\}}@*)  
Reasoning Question:  
(*@\texttt{\{EX\_QUESTION\_1\}}@*)  
Answer:  
(*@\textbf{\{EX\_ANSWER\_1\}}@*)

---

(*@\textbf{Example 2}@*)  
Task: (*@\texttt{\{EX\_TASK\_2\}}@*)  
Patient Data (CSV):  
(*@\texttt{\{EX\_PATIENT\_DF\_2\}}@*)  
Vital Sign Image:  
(*@\texttt{\{EX\_IMAGE\_2\}}@*)  
Reasoning Question:  
(*@\texttt{\{EX\_QUESTION\_2\}}@*)  
Answer:  
(*@\textbf{\{EX\_ANSWER\_2\}}@*)

---

(*@\textbf{Now analyze the following patient case}@*)  

Task: (*@\texttt{\{TASK\}}@*)  
Patient Data (CSV):  
(*@\texttt{\{PATIENT\_DF\}}@*)  
Vital Sign Image:  
(*@\texttt{\{IMAGE\}}@*)  
Reasoning Question:  
(*@\texttt{\{QUESTION\}}@*)

---

Instructions:
- Analyze trends, values, and correlations from both the CSV and the image
- Do not use external medical knowledge
- Return only: `1` or `0`
- (*@\textbf{No explanations. No extra text. Strictly one digit.}@*)
\end{lstlisting}
\noindent\textbf{N. CoT prompt for Reasoning-QA evaluation based on Text, Time Series and Image:}

\begin{lstlisting}[caption={Multimodal ICU Binary Reasoning Prompt}]
You are a medical expert in interpreting ICU time-series physiological data and trend images.

You are given:
- ICU patient data in CSV format
- A plot of time-series trends for:
  - HR (Heart Rate)
  - DiasABP (Diastolic Arterial BP)
  - SysABP (Systolic Arterial BP)
  - RespRate (Respiratory Rate)
  - Temp (Body Temperature)
  - Glucose (Blood Glucose Level)

Your task is to:
1. Examine the data and visual trends
2. Decide whether the question can be definitively answered
3. Return `1` if the data supports a clear answer, or `0` if insufficient/inconclusive

---

(*@\textbf{Example 1}@*)  
Task: (*@\texttt{\{EX\_TASK\_1\}}@*)  
Patient Data (CSV):  
(*@\texttt{\{EX\_PATIENT\_DF\_1\}}@*)  
Vital Sign Image:  
(*@\texttt{\{EX\_IMAGE\_1\}}@*)  
Reasoning Question:  
(*@\texttt{\{EX\_QUESTION\_1\}}@*)  

Step-by-Step Reasoning:  
(*@\texttt{\{EX\_REASONING\_1\}}@*)  

Answer:  
(*@\textbf{\{EX\_ANSWER\_1\}}@*)

---

(*@\textbf{Example 2}@*)  
Task: (*@\texttt{\{EX\_TASK\_2\}}@*)  
Patient Data (CSV):  
(*@\texttt{\{EX\_PATIENT\_DF\_2\}}@*)  
Vital Sign Image:  
(*@\texttt{\{EX\_IMAGE\_2\}}@*)  
Reasoning Question:  
(*@\texttt{\{EX\_QUESTION\_2\}}@*)  

Step-by-Step Reasoning:  
(*@\texttt{\{EX\_REASONING\_2\}}@*)  

Answer:  
(*@\textbf{\{EX\_ANSWER\_2\}}@*)

---

(*@\textbf{Now analyze the following patient case using the same reasoning process}@*)

Task: (*@\texttt{\{TASK\}}@*)  
Patient Data (CSV):  
(*@\texttt{\{PATIENT\_DF\}}@*)  
Vital Sign Image:  
(*@\texttt{\{IMAGE\}}@*)  
Reasoning Question:  
(*@\texttt{\{QUESTION\}}@*)

---

Step-by-Step Reasoning:  
(Identify and correlate physiological patterns from data and plot. Determine if they support a clear conclusion.)

Final Answer:  
(*@\textbf{Return only one digit: 1 or 0}@*)
\end{lstlisting}

\twocolumn

\bibliographystyle{unsrtnat}
\bibliography{template}  






\end{document}